\documentclass[conference]{IEEEtran}
\usepackage{times}

\usepackage[numbers]{natbib}
\usepackage{multicol}
\usepackage[bookmarks=true]{hyperref}
\usepackage{graphicx}
\usepackage{lipsum}
\usepackage{amsmath}
\usepackage{amssymb}
\usepackage[dvipsnames]{xcolor}

\usepackage{mathtools, algorithm, algpseudocode}
\usepackage{cleveref}

\definecolor{mypink}{rgb}{0.858, 0.188, 0.478}
\definecolor{mygold}{rgb}{0.890196078, 0.631372549, 0.176470588}

\definecolor{hiroshige}{HTML}{ffd06f}

\definecolor{darkBlue}{RGB}{10,50,220}
\definecolor{customGreen}{RGB}{112,173,71}

\definecolor{rand}{HTML}{2CA02C}
\definecolor{eig}{HTML}{9467BD}
\definecolor{cost_eig}{HTML}{D62728}
\definecolor{rand_task}{HTML}{8C564B}
\definecolor{task_eig}{HTML}{FF7F0E}
\definecolor{cost_task_eig}{HTML}{1F77B4}

\definecolor{optimal}{HTML}{36ADA4}
\definecolor{verb}{HTML}{97A431}
\definecolor{lang}{HTML}{F77189}
\definecolor{random}{HTML}{A48CF4}

\newcommand{\truedist}{P_{ij}}
\newcommand{\learneddist}{Q_{ij}}
\newcommand{\trial}{x_{ij}}
\newcommand{\sample}{\hat{x}_{ij}}

\hypersetup{
	colorlinks=true,
	linkcolor=darkBlue,
	filecolor=magenta,      
	urlcolor=cyan,
	citecolor=magenta,
}

\begin{document}

\title{Efficient Evaluation of Multi-Task \\Robot Policies With Active Experiment Selection}
% You will get a Paper-ID when submitting a pdf file to the conference system
% \author{Author Names Omitted for Anonymous Review. Paper-ID 708}

\author{\authorblockN{Abrar Anwar, Rohan Gupta, Zain Merchant, Sayan Ghosh, Willie Neiswanger, Jesse Thomason}
\authorblockA{University of Southern California \\ Contact: \{\texttt{abrar.anwar}, \texttt{jessetho}\}\texttt{@usc.edu}}
}
% \and
% \authorblockN{Homer Simpson}
% \authorblockA{Twentieth Century Fox\\
% Email: homer@thesimpsons.com}
% \and
% \authorblockN{James Kirk\\ and Montgomery Scott}
% \authorblockA{Starfleet Academy\\
% San Francisco, California 96678-2391\\
% Telephone: (800) 555--1212\\

% avoiding spaces at the end of the author lines is not a problem with
% conference papers because we don't use \thanks or \IEEEmembership

%Homer Simpson\authorrefmark{2},
%Montgomery Scott\authorrefmark{3} and
%Eldon Tyrell\authorrefmark{4}}
% \authorblockA{\authorrefmark{1}School of Electrical and Computer Engineering\\
% Georgia Institute of Technology,
%Telephone: (800) 555--1212, Fax: (888) 555--1212}
%\authorblockA{\authorrefmark{4}Tyrell Inc., 123 Replicant Street, Los Angeles, California 90210--4321}}

\maketitle

\begin{abstract}
% As robots become increasingly capable at solving many tasks, evaluating these robot policies in the real world is time-consuming for experimenters.
Evaluating learned robot control policies to determine their physical task-level capabilities costs experimenter time and effort.
The growing number of policies and tasks exacerbates this issue.
It is impractical to test every policy on every task multiple times; each trial requires a manual environment reset, and each task change involves re-arranging objects or even changing robots.
Naively selecting a random subset of tasks and policies to evaluate is a high-cost solution with unreliable, incomplete results.
In this work, we formulate robot evaluation as an active testing problem.
We propose to model the distribution of robot performance across all tasks and policies as we \textit{sequentially} execute experiments.
Tasks often share similarities that can reveal potential relationships in policy behavior, and we show that natural language is a useful prior in modeling these relationships between tasks.
% as an experimenter evaluates robot policies.
We then leverage this formulation to reduce the experimenter effort by using a cost-aware expected information gain heuristic to efficiently select informative trials.
Our framework accommodates both continuous and discrete performance outcomes.
We conduct experiments on existing evaluation data from real robots and simulations.
By prioritizing informative trials, our framework reduces the cost of calculating evaluation metrics for robot policies across many tasks. 
% By reducing cost of experiment sampling for robotics, we hope that our framework enhances the efficiency of robot policy testing.

\end{abstract}

\IEEEpeerreviewmaketitle

\section{Introduction}

% TODO: depending on how well language helps, we can maybe fron tthat intuition at the beginning of the paper.
With the growth of large-scale robot datasets and pretrained policies, robot systems have become capable of achieving good performance across many tasks; however, this diversity makes evaluating these policies increasingly more difficult.
% As these robots become more capable across a large number of tasks, the evaluating these robot policies become difficult, as these policies are trained on large datasets with dozens of tasks. 
% However, as robots become increasingly proficient in handling a diverse set of tasks, evaluating these policies becomes increasingly complex. 
Unlike fields such as computer vision or natural language processing, physical robotics experiments are conducted sequentially, with each policy rollout taking experimenter effort.
Considering the effort to change task setups, it becomes impractical to evaluate every policy on every task.

In practice, experimenters are typically interested in selecting the best checkpoints, tuning hyperparameters, or comparing model architectures, which do not necessarily require a full evaluation across every policy across every task.
A robot policy that can ``pick up an apple" is likely capable of ``picking up an orange" for an analogous scene.
Our insight is to take advantage of relationships between tasks and frame evaluation as a population parameter estimation problem, which lets us design more efficient experiment sampling strategies.
% We can then use these relationships to efficiently select experiments that have the least cost to 

% Roboots are getting getting more capable in the tasks they can do.
% Octo claims XYZ tasks, OpenVLA claims XYZ tasks. The datasets have XYZ tasks.
Manipulation~\cite{octo_2023, kim24openvla, black2024pi_0} and navigation~\cite{shah2023lm,shah2023vint,anwar2024remembr} approaches continue to improve.
% constantly being improved upon; however, as new methods become increasingly capable in accomplishing tasks, the evaluation of all these methods across tasks become difficult.
Simulation-based evaluation has become a common approach to measure that improvement~\cite{simpler_env}, but simulation has often been insufficient for understanding real-world performance~\cite{anderson2021sim,deitke2020robothor,simpler_env}.
% While simulation has become a common approach to evaluating language-guided policies ~\cite{simpler_env, shridhar2020alfred, mees2022calvin, anderson2018vision}, it is often insufficient for understanding real-world performance~\cite{anderson2021sim, deitke2020robothor, simpler_env}.
% However, evaluating robots in the physical world is challenging due to this large domain of tasks in their training set and the need to test multiple policies. 
The combinatorial growth of tasks with scene complexity makes an exhaustive evaluation even more impractical.
As such, there is a need for efficient evaluation strategies that can enable systematic and scalable testing of multi-task robot policies in the real world.

% In practice, experimenters are often interested in identifying the best and worst policies, which can help in selecting checkpoints, tuning hyperparameters, or comparing model architectures.
% In this work, we propose an efficient evaluation strategy that takes advantage of relationships between tasks and policies by framing the problem as a matrix completion problem.
\begin{figure}[t]
    \centering
    \includegraphics[width=1\linewidth]{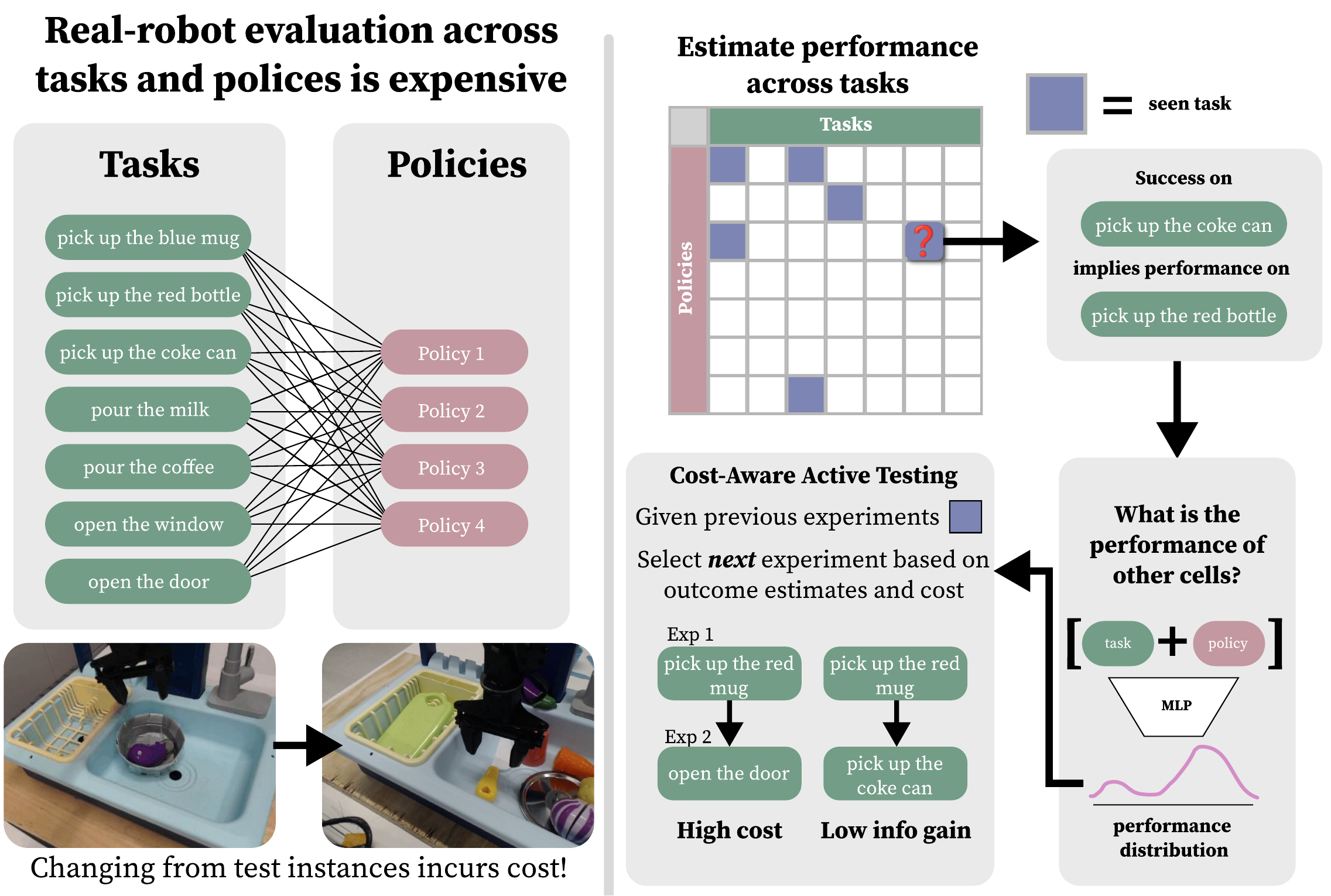}
    \caption{\textbf{Overview.} 
    Exhaustively evaluating multiple robot policies across various tasks has high experimenter cost.
    In this work, we leverage latent relationships between tasks and policies to model performance distributions across all tasks and policies. 
    These estimates are updated sequentially and used to implement cost-aware active experiment selection strategies.
    }
    \label{fig:teaser}
\end{figure}
% Novelty 1. We learn a surrogate model to estimate the distribution of parameters
When evaluating a robot policy, it is common to consider only the mean of some metric.
However, since robot performance often has high variance, we instead consider the evaluation of a policy on a specific task as a distribution of outcomes.
% Instead, we consider the evaluation of a policy on a specific task as a distribution of outcomes, as 
% Thus, every policy-task pair is defined by a random variable with parameters such as $\mu, \sigma$ for a Gaussian distribution.
Thus, every policy-task pair is characterized by a distribution reflecting the experiment conditions, for example a Bernoulli distribution for binary success or a Gaussian distribution for a reward outcome.
In this work, as an experimenter conducts evaluations sequentially, we learn a surrogate model that estimates the parameters for this distribution for every policy-task pair under consideration.

% Novelty 2. We show that across tasks, there is shared information that can be leveraged to better estimate the parameters of the distribution.
To build an efficient evaluation strategy, we take advantage of latent shared structure between tasks. 
% Intuitively, a robot that can ``pick up an apple" is likely capable of ``picking up an orange".
% Thus, it is \textit{likely} that an experimenter would not have to evaluate both tasks, which would save effort.
As we sample new experiments, we learn a surrogate model conditioned on latent task and policy embeddings. 
We show that better representations of a policy and a task, including language-based priors for tasks, improves estimates of the outcome distributions, indicating that there is shared information between tasks and policies learnable from policy performance.

% Novelty 3. We can use the surrogate model to actively sample experiments in a way that is cost effective for the experimenter.
Since evaluation is expensive, we want to minimize the cost of evaluation while still estimating the performance of all policies across all tasks of interest.
Then, with our surrogate model, we leverage strategies from the active learning literature to integrate cost-efficient sampling heuristics like expected information gain.
We show that our approach is able to efficiently estimate the performance of robot policies across tasks.

In particular, we:
\begin{itemize}
% \begin{itemize}[nosep,leftmargin=*] %[noitemsep,topsep=0pt]
    % \item We propose several key desiderata for VLN evaluation methods that encourage targeted evaluation of the linguistic and visual capabilities of physical ground robots without the need for dramatic physical costs required to make simulations-sized test sets
    \item formalize multi-task robot policy evaluation as a population parameter estimation problem;
    \item find that there are performance relationships between tasks for estimating the performance of a policy-task pair;
    \item create an active testing protocol that leverages these performance relationships between tasks and policies, allowing us to efficiently evaluate multiple robot policies;
    \item and create cost-aware sampling strategies that can estimate the performance of robot policies with lower cost.
    % \item design an approach to estimate the performance of robot policies on unseen test instances;
    % \item leverage the low-rank structure of robot policy outcomes to efficiently estimate the performance of unseen robot policies by learning task and policy embeddings over time;
\end{itemize}

\section{Background and Related Work}
% We methods used for evaluating machine learning models, methods for evaluating robot policies, and introduce the concept of contrast sets.

% Past work in machine learning model evaluation has used perturbations as a method to probe model performance; however, evaluation in robotics typically focuses on a small number of pre-defined tasks. In this section, we discuss contrast sets from NLP, and its relevance to evaluation for robotics.
Past work in machine learning model evaluation and active learning have considered how to compare model performance; however, as more robot policies become easier to develop, it is critical to develop better strategies for evaluating robot strategies.
We discuss approaches for testing models in machine learning and its relevance to evaluation for robotics.

\textbf{Evaluation in Machine Learning.}
% \textcolor{red}{TODO: cite sriram's lab's paper, LM eval papers, etc. differentiate ourseles from those}
% As machine learning models become more prevalent in real-world settings, research has begun to shift towards evaluating these models.
In fields such as computer vision or NLP, it is common to characterize the out-of-distribution performance of a single model~\cite{wang2021cline, hendrycks2020pretrained, recht2019imagenet, hendrycks2019benchmarking,liang2022holistic, chang2024survey, gardner2020evaluating}, some of which create a standard for comparing different models as well~\cite{liang2022holistic}. 
These approaches allow for experimenters to quickly understand the performance of their model, and in some cases compare between models.
However, in robotics, each task is expensive to evaluate and each policy evaluation is difficult. 
In this work, we look at use methods from active learning to improve experiment selection during evaluation.
% A large, sampled i.i.d. test set may not capture the span of expected situations a machine learning model could encounter in the real world. 
% To address this, researchers have designed out-of-distribution evaluation techniques in the vision~\cite{wang2021cline, hendrycks2020pretrained, recht2019imagenet, hendrycks2019benchmarking} and NLP~\cite{liang2022holistic, chang2024survey, gardner2020evaluating} communities. 
% In computer vision, perturbations of images have been used to generate counterfactual examples to test a model~\cite{zemni2023octet, jeanneret2023adversarial, jeanneret2022diffusion, sauer2021counterfactual, prabhu2024lance}.
% In NLP, contrast sets~\cite{gardner2020evaluating} perturb the original test set to accurately evaluate a model's true linguistic capabilities.
% These approaches allow experimenters to stress test their models and have confidence in their model during deployment.
% However, in robotics, testing requires physical deployment and takes considerable efforts to compute such metrics.
% In this work, we focus on designing contrast sets for language-guided robot policies.

\textbf{Active Testing.}
% \lipsum[2]
Similar to active learning which aims to select training labels, active testing approaches~\cite{sawade2010active, rainforth2024modern, yilmaz2021sample} focus on selecting test instances to evaluate to better predict model performance. 
Though these settings focus on classification or regression labeling tasks, this formulation is important to robotics as evaluation is expensive.
Various Bayesian optimization, active learning, and active testing approaches use surrogate models to estimate the value of a training or test instance~\cite{eggensperger2015efficient, brochu2010tutorial, shahriari2015taking, cozad2014learning,qian2006building, kossen2021active}, often incorporating cost-aware sampling~\cite{lee2020cost, paria2020cost}.
In robotics, surrogate models have been used to predict outcomes of a human-robot interaction scenarios in simulation for policy learning~\cite{bhatt2023surrogate}; however, that past work did not consider the cost evaluating each scenario.
Additionally, most of these works focus on active learning and active testing for regression models.
Since robot evaluation can have high variance, we take inspiration from past work~\cite{tosh2022targeted} to focus on active learning of probablistic models using a surrogate model.
We then apply these cost-aware active testing strategies on multi-task, multi-policy robot evaluation by learning a task and policy conditioned surrogate model.
% In this work, we build on active testing approaches to actively select experiments that are most informative for learning a surrogate model, and we show that this surrogate model is able to accurately estimate the multi-task, multi-policy performance distributions.

\textbf{Evaluation of Robot Policies.}
% Sim2real transfer strategies encompass various approaches, often employing domain randomization~\cite{tobin2017domain} or using generative adversarial networks to shift the target domain observation closer to the source domain~\cite{rao2020rl}. 
% However, this work's focus is not on the actual sim2real transfer process.
% Instead, our aim is to evaluate policies more effectively in the real world.
% Simulation is a common way to train and evaluate robot policies~\cite{nasiriany2024robocasa, mandlekar2021matters, james2020rlbench}. 
Simulation is often used to evaluate the performance of a real-robot system \cite{deitke2020robothor, anderson2021sim, kadian2020sim2real, gervet2023navigating} by recreating a simulated counterpart to a real environment, but shows ineffective direct sim2real performance without domain randomization or real-world finetuning strategies.
There exist correlations between simulation and real-world performance even if they do not exactly match~\cite{wilbert_colloseum, simpler_env}; however there are no guarantees about real-world performance.
% These works also pre-define a set number of tasks in simulation; but it is not scalable to engineer simulators for every new task.
Other recent work focuses on real world evaluation such as carefully selecting the initial conditions of an experiment~\cite{kress2024robot}, evaluating LLM-based task planners~\cite{hu2024deploying}, active capability assessment of black-box symbolic planners~\cite{verma2021discovering, verma2023autonomous, nayyar2022differential}, or providing bounds on policy performance by assuming some underlying distribution for outcomes~\cite{that_tri_paper}.
Other work has investigated how changes to these initial conditions can provide information about policy sensitivity~\cite{parekh2024investigating,xie2024decomposing,anwar2024contrast} or has used factors of the initial conditions and naive sampling strategies to more efficiently collect data~\cite{gao2024}.
In this work, we consider the setting of evaluating multiple policies across various tasks while also learning the parameters of an underlying distribution. 
We then leverage this learned distribution to more efficiently sample experiments for evaluation.

\section{Problem Formulation and Notation}

\begin{figure*}[ht]
    \centering
    \includegraphics[width=\linewidth]{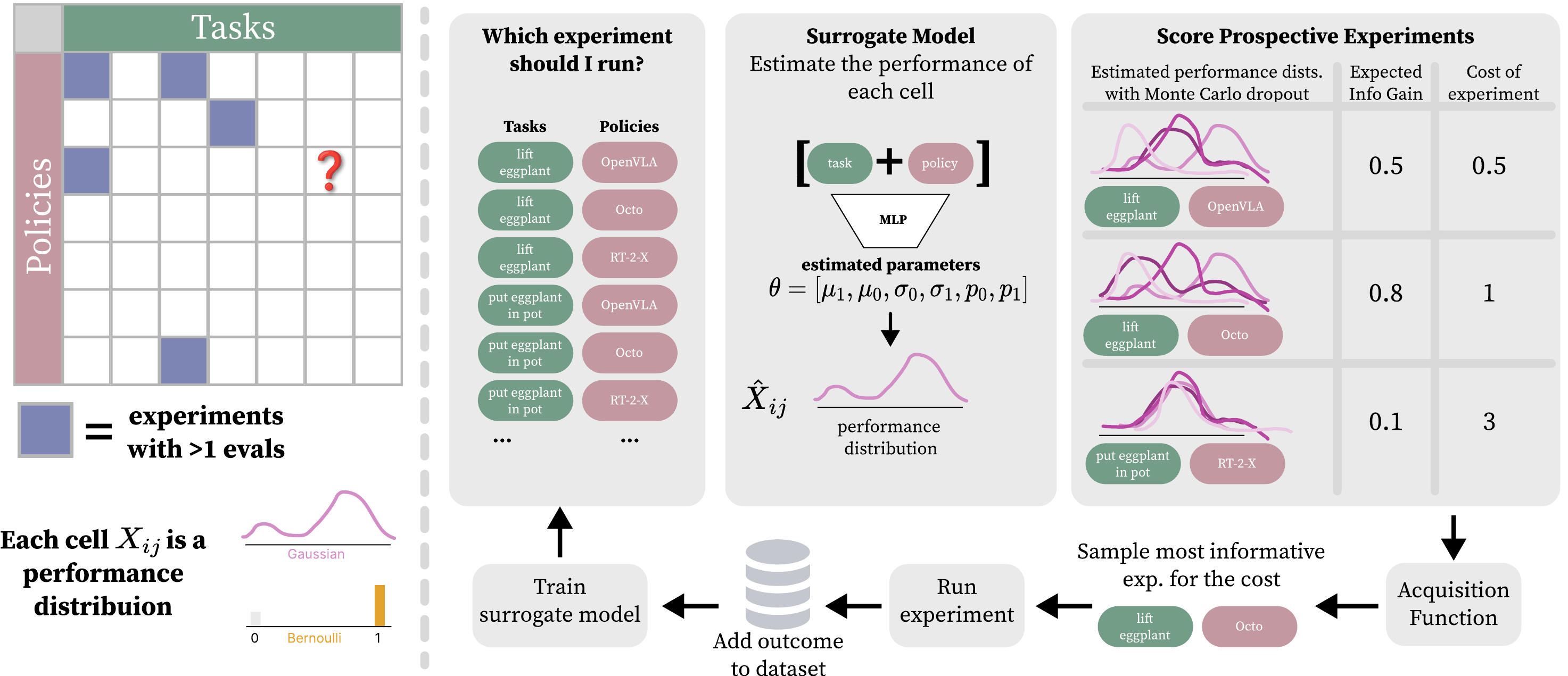}
    \caption{\textbf{Method.}
    % When an experimenter has multiple robot policies across various tasks, it becomes difficult to evaluate a robot across all these combinations.
    % In this work, we note that likely have relationships between each other that may indicate a task's performance. 
    We build a surrogate parameter estimation model that learns task and policy embeddings to predict the outcome performance distribution of a task and policy combination. 
    We use Bernoulli distributions for binary outcomes or a bimodal Gaussian for continuous outcomes.
    Given this parameter estimation model, we develop an active testing strategy with cost-aware sampling based on expected information gain.
    }
    \label{fig:overview}
\end{figure*}

% In this work, we address the challenge of efficiently evaluating multi-task robot policies in the real world.
The objective of this work is to design an efficient strategy to evaluate robot policies across tasks while balancing the cost of experimentation. 
Consider a fixed set of $M$ robot policies, denoted by $\mathcal{P} = \{\pi_1, \pi_2, \ldots, \pi_M\}$ and a set of $N$ tasks $\mathcal{T} = \{T_1, T_2, ..., T_N\}$. 
% Due to practical constraints, only a subset of $N$ tasks $A \subset \mathcal{T}$ is used.
Each task $T_j \in \mathcal{T}$ is a finite-horizon MDP defined by states, actions, and a high-level natural language instruction $L_i$.
% Since robot policies can be sensitive to changes in language instructions~\cite{}, we assume a one-to-one mapping between language instructions and tasks.

Our framework is policy-agnostic and does not assume access to policy model weights, and can be applied to engineered robot \textit{systems} in addition to end-to-end models.

\textbf{Population Parameter Estimation}. 
We formulate the problem as population parameter estimation, similar to probabilistic matrix factorization~\cite{mnih2007probabilistic}.
Let the performance of a policy ${\pi_i \in \mathcal{P}}$ on a task $T_j \in \mathcal{T}$ be represented by the random variable $X_{ij}$ with distribution $P_{ij}$, from which we can sample evaluations $\trial \sim \truedist$. Here, $\truedist$ represents the ``true'' performance distribution.
Since the underlying distribution $\truedist$ is unknown, the goal of population parameter estimation is to estimate a distribution $\learneddist$ that models real-world evaluation outcomes from $\truedist$.
% , which is the distribution corresponding to a random variable $\hat{X}_{ij}$, 
We use $\theta_{ij}$ to represent the parameters of the learned distribution $\learneddist$.
For example, $\theta_{ij}=[\mu, \sigma]$ if $\learneddist$ is a Gaussian distribution.
Given a limited number of observed samples from the true distribution, $\trial^1, ..., \trial^n \sim \truedist$, the goal is to estimate the parameters of an estimated distribution $\theta_{ij}$.
Our setting also has samples from other random variables, $X_{kl}$ corresponding to different policy-task pairs. 
Therefore, in this work we want to estimate $\Theta = \{\theta_{ij}\}_{i, j = 1}^{i=M, j = N}$ for all policy-task pairs given a dataset $\mathcal{D}=\{\trial^k\}$.
These distributions can be visualized as a grid of policy-task pairs as shown in Figure~\ref{fig:overview}.

The aim is to estimate the parameters of $\learneddist$ of all policy-task combinations by leveraging shared information across this matrix.
However, it is infeasible to directly evaluate all policy-task pairs due to cost constraints. 
Therefore, we adopt an active testing approach, where the objective is to iteratively select the most informative experiments $(\pi_i, T_j)$ to efficiently learn $\Theta$. 
% Given a limited evaluation budget, we develop active experiment selection strategies to efficiently samples evaluations that improve the accuracy of these parameter estimates.

\textbf{Active Testing.}
% Since it is difficult to sample a large set of outcomes $\mathcal{D}$ due to experimenter effort, we take an active testing approach.
We apply an active learning paradigm to learn a population parameter estimator $f(\pi_i, T_j)$.
As such, we define acquisition functions to guide the selection of task-policy pairs or tasks alone, and then sample experiments that are most informative.
First, we define an acquisition function $a(\pi_i, T_j)$, and the next experiment is selected by maximizing this function over all possible experiments:
\begin{equation}
    (\pi_i^*, T_j^*) = \arg\max_{(\pi_i, T_j)} a(\pi_i, T_j).
\end{equation}
Although these acquisition functions are informative, we want a balance between selecting informative experiments and their costs.

\textbf{Evaluation Cost}. In real-world evaluation, each policy-task evaluation incurs a cost. 
Let $c_{\text{eval}}(T_j)$ denote the cost of a single evaluation of a policy on task $T_j$.
We make a simplifying assumption that this cost is agnostic to changes in the policy under evaluation, that often being a configurable software option.
This cost could include the time required to execute the policy, the resources consumed during evaluation, or the manual supervision required to reset the scene.
Furthermore, switching between tasks typically incurs a larger cost involving a reconfiguring the scene or the robot. 
We define this switching cost $c_{\text{switch}}(T_j, T_k)$ as the cost associated with transitioning from task $T_j$ to $T_k$.
% This cost depends on the specific tasks being evaluated.
For a sequence of tasks that have been evaluated $T_{i_1}, \ldots, T_{i_L}$ (where each $i_j \in N$), we compute the total cost of evaluation as:

% Do we need this part? wrote it like this to take up more space
$$c_{\text{total}} = \sum_{j=1}^N c_{\text{eval}}(T_{i_j}) + \sum_{j=1}^{N-1} c_{\text{switch}}(T_{i_j}, T_{i_{j+1}})$$

Given these costs, the problem is to design an evaluation strategy that minimizes the total cost of evaluation while learning the population parameters of test instances.

\section{Method}

We aim to design a framework for sampling experiments for multi-task robot policies.
Our framework consists of two parts: (1) learning a surrogate model to estimate the population parameters of a test instance and (2) designing strategies to sample experiments in a cost-efficient manner.
The surrogate model leverages task and policy representations that define an experiment to have a better estimate of the overall performance distributions.
Then, we use this surrogate model to compute the expected information gain of different experiments.
We then use the expected information gain along with the cost of switching tasks to conduct active testing.

% \textbf{Surrogate Model.}
\subsection{Surrogate Model}
As we evaluate our robot policies across tasks, we track the outcomes of each trial to aggregate a dataset $\mathcal{D}$ over time.
% In the process of evaluating multi-task robot policies, we collect a dataset of previous outcomes from trials $\trial \in \mathcal{D}$.
Each of these outcomes are realizations of a true underlying distribution $\truedist$.
Our goal is to learn a surrogate model from $\mathcal{D}$ that predicts the population parameters $\theta_{ij}$ of a performance distribution  $\learneddist$. 
As more evaluation rollouts are conducted, we add the outcomes to $\mathcal{D}$ and continue training the surrogate model.
% A single experiment can be defined by the policy $\pi_i$ being executed on task $t_j$.

%The surrogate model should be able to capture the performance relationship between tasks and policies.
To train an effective surrogate model,  we use notions of similarity between tasks and policies. 
% For instance, tasks that involve similar skills, such as "picking up an apple" and "picking up an orange," should result in similar performance distributions for a given robot policy.
Thus, we need a representation that captures the similarities between policies and tasks with respect to their performance distributions.
We define a policy embedding $e_{\pi_i}$ and task embedding $e_{T_j}$, where similar performance distributions in task and policy can be captured based on the embeddings.
These policy and task representations are then provided as input to an MLP that predicts the estimated population parameters:
\begin{equation}
    \hat{\theta}_{ij} = f(\pi_i, T_j) = \text{MLP}(e_{\pi_i}, e_{T_j}).
\end{equation}

\textbf{Task and Policy Representation.} 
\label{sec:method}
To define the task and policy embeddings $e_{\pi_i}, e_{T_j}$, we design various types of embeddings.
In practice, we cannot know the relationship between policies in advance as we are conducting evaluation.
Therefore, we define the policy embedding to be a fixed, randomly initialized embedding to act as an identifier for the policy in a given experiment.

For the task embedding $e_{\pi_i}$, we leverage language embeddings from MiniLMv2~\cite{gu2024minillm} which we reduce to 32 dimensions using PCA over all tasks.
However, we found that language embeddings overly focus on nouns as opposed to verbs, which causes issues as actions with similar nouns but different verbs would be closer together verbs with the same nouns. 
Thus, we apply the following procedure to mitigate this issue.
We (1) use part-of-speech tagging to extract all verbs and verb phrases, (2) compute a language embedding for the verb $e_{T_j}^{\text{verb}}$ and for the entire task description  $e_{t_j}^{\text{task}}$, and then (3) compute the task embedding 
\begin{equation}
     e_{T_j} =  0.8 \cdot e_{T_j}^{\text{verb}} +  0.2\cdot e_{T_j}^{\text{task}} + 0.1\cdot \mathcal{N}(0,1).
\end{equation}
We also found that the embeddings were often too close across multiple tasks, and we found that adding a slight noise term helped separate close embeddings.
Experiments on this result are in Section~\ref{sec:task_results}.
% These embeddings were them 

\textbf{Population Parameter Estimation.}
Outcomes in robot learning can take the form of continuous values like rewards, time to completion, or task progress, and binary values like task success.
Thus, the underlying distribution from the surrogate model depends on the type of task.
We consider two types of underlying distributions.
When $X_{ij}$ is continuous, $\learneddist$ takes the form of a mixture of Gaussians with $K$ components,
\begin{equation}
    % P(X_{ij}; \theta_{ij}) = \sum_{k=1}^{K}\pi_k\mathcal{N}(\mu_k, \sigma_k),\\
    \sample \sim \learneddist = \sum_{k=1}^{K}p_k\mathcal{N}(\mu_k, \sigma_k),
\end{equation}
where $\pi_k, \mu_k,$ and $\sigma_k$ are the mixing coefficients, means, and standard deviations of the Gaussian components respectively that are predicted from the surrogate model $\theta_{ij} = f(\pi_i, T_j)$.
% The parameters $\theta_{ij}=\{\pi_k, \mu_k, \sigma_k\}^{K}_{k=1}$ are predicted using the surrogate model.
We thus train the surrogate model with a mixture density loss~\cite{bishop1994mixture,ha2018world} to minimize the negative log-likelihood of the observed data under the mixture model.
In our experiments on continuous outcome distributions, we use $K=2$ Gaussian components, as robotics performance is often bimodal; robots either fail catastrophically or they maintain non-zero performance.

In the case where $X_{ij}$ is binary, indicating success or failure, $\learneddist$ takes the form of a Bernoulli distribution:
\begin{equation}
    % P(X_{ij}; \theta_{ij}) = p^{x_{ij}}(1-p)^{1-x_{ij}}
    \sample \sim \learneddist =  p^{x_{ij}}(1-p)^{1-x_{ij}},
\end{equation}
where $\theta_{ij} = \{p \in [0,1] \}$ represents the success probability predicted by the surrogate model trained using cross-entropy loss.

\subsection{Cost-aware Active Experiment Selection}

% We can now use this surrogate model for selecting policy-task robot experience to execute. 
We explore cost-aware, active-experiment acquisition functions that guide selection of experiments based on their expected utility while considering associated costs.
To define the acquisition function, we first focus on how to measure the informativeness of a policy-task evaluation, which we capture through expected information gain.

\textbf{Expected Information Gain.}
Expected Information Gain (EIG) quantifies the value of an experiment by estimating how much it reduces the predictive uncertainty of the performance distribution for a policy-task pair. 
Since the surrogate model estimates performance \textit{distributions}, we define the EIG of a policy-task pair using a Bayesian Active Learning by Disagreement (BALD)~\cite{houlsby2011bayesian} formulation for probabilistic models
\begin{equation}
    \mathcal{I}(\pi_i, T_j) = \underbrace{\mathbb{H}[\learneddist]}_{\text{marginal entropy}} - \underbrace{\mathbb{E}_{\theta_{ij} \sim f(\theta_{ij}|\mathcal{D})} [\mathbb{H}[\learneddist | \theta_{ij}]]}_{\text{expected conditional entropy}}.
\end{equation}
% The first term is the entropy over the marginal predictive distribution
The first term represents the marginal entropy over $Q_{ij}$, which quantifies the total uncertainty in $Q_{ij}$. 
The second term corresponds to the expected conditional entropy over multiple samples of parameters $\theta_{ij}$.
% where $P(\learneddist|\mathcal{D})$ is the marginal predictive distribution over the performance distribution $\learneddist,$ while $P(\learneddist | \theta_{ij})$ is the conditional predictive distribution given a sampled set of parameters $\theta_{ij}$. 
% Thus, the first term is the entropy over the marginal predictive distribution and captures the epistemic and aleatoric uncertainties of the model's predictions, while the second term captures the expected aleatoric uncertainty.
Thus, $\mathcal{I}(\pi_i, T_j)$ captures the disagreement between multiple samples of distributions.
For example, if 10 sampled parameters for a Gaussian have very different distributions, then their disagreement will be high.
Since the entropy of a mixture of Gaussians generally lacks a closed-form solution, we estimate the entropy by discretizing the empirical distribution into $n=25$ bins for which to compute entropy over.

BALD ensures the EIG score is higher in test instances where there is disagreement in the predicted distributions across sampled parameters.
In this case, we define the acquisition functions $a(\pi_i,T_j)=\mathcal{I}(\pi_i,T_j)$.

To compute the expected information gain, we require multiple samples of $\Theta_{ij}$; however, we only train a single MLP.
Inspired by Monte Carlo dropout~\cite{gal2016dropout} and past literature~\cite{loquercio2020general,ledda2023dropout}, we apply dropout only at test-time to compute multiple samples of $\theta_{ij}$ from the surrogate model $f(\cdot)$.
% We found that dropout early in the sampling process would lead to overfitting on the small dataset, leading to low entropy in the samples of $\Theta_{ij}$.

% We use Monte Carlo (MC) dropout~\cite{gal2016dropout} to estimate sampling from $f(\Theta_{i,j} | \mathcal{D})$. 
% We found that Monte Carlo (MC) dropout~\cite{gal2016dropout}, where dropout is applied during training time, then at test time to compute multiple samples of $\theta_{ij}$ from the surrogate model $f(\cdot)$, leads to overfitting on the small amounts of training data available early in the evaluation process. 
% This overfitting lead to the disagreement between sampled parameters to be nearly zero, causing the EIG scores to be useful after hundreds of expensive evaluations.

% However, in the setting of robot evaluation where experiments are expensive, our surrogate model is initially trained with minimal data.
% We found that dropout during training time early in the evaluation process led to overfitting, as few examples had been seen, which caused the disagreement between sampled parameters to be essentially zero.
% Thus, the EIG scores would only be useful after hundreds of expensive evaluations.
% Past work in the active learning literature~\cite{tosh2022targeted} avoids this cold start problem by initializing their model with hundreds or thousands of training instances, which is impractical for an evaluation framework.
% To address this cold start problem, we found that restricting dropout to test time only effectively allowed our model to provide useful EIG scores earlier in the evaluation process.

\textbf{Cost-Aware EIG}.
While EIG effectively quantifies the informativeness of an experiment, it does not consider the costs of conducting evaluation.
To make EIG cost-aware, we design the following acquisition function based on prior work that simply integrates cost with a multiplicative factor~\cite{paria2020cost,lee2020cost}:
\begin{equation}
    a_{\text{cost-aware}}(\pi_i, T_j, T_\text{current}) = \dfrac{\mathcal{I}(\pi_i, T_j)}{(\lambda \cdot c_{\text{switch}}(T_{\text{current}}, T_j))+1},
    \label{eq:cost_aware_eig}
\end{equation}
where $\mathcal{I}(\pi_i, T_j)$ represents EIG for the policy $\pi_i$ on task $T_j$, $c_{\text{switch}}(T_{\text{current}}, T_j))$ is the cost of switching from the current task $T_{\text{current}}$ to a new task $T_j$, and $\lambda$ is a hyperparameter that controls the cost sensitivity.

\begin{algorithm}
\caption{Active Experiment Selection Procedure}
\label{alg:short_pseudocode}
\begin{algorithmic}[1]
\Require A set of policies $\pi_i \in \mathcal{P}$ to evaluate over tasks $T_j \in \mathcal{T}$, an empty dataset of outcomes $\mathcal{D}$, an untrained surrogate model $p(\pi_i, T_j)$, exploration rate $\epsilon = 0.1$
\State Randomly sample a single task $T_j$ and evaluate every policy 3 times. Add outcomes $x_{ij}^k$ to $\mathcal{D}$
\State Set $T_{\text{current}} = T_j$
\State Increment $C_\text{total} = C_\text{eval} + c_\text{eval} \cdot |\mathcal{P}| \cdot 3$ 
\State Train the surrogate model $p(\cdot)$ on $\mathcal{D}$ for $k$ epochs
\For{each query step}
    \State Use MC dropout to sample 10 predicted distributions from the surrogate model for every policy-task pair
    \State Use sampled distributions to compute scores $s_{ij}=a(\pi_i, T_j, T_\text{current})$ according to Eq.~\ref{eq:cost_aware_eig}
    \State With probability $\epsilon$, select a random $(\pi_i, T_j)$
    \State Otherwise, select $(\pi_i, T_j) = \arg\max_{(\pi_i, T_j)} s_{ij}$

    \State Conduct 3 evaluations and observe  $\trial^1, \trial^2, \trial^3 \sim \truedist$
    \State Add these outcomes to $\mathcal{D}$
    \State Train $f(\cdot)$ on $\mathcal{D}$ for $k$ epochs

    \State Increment $C_\text{total} = C_\text{total} + c_\text{eval} \cdot 3$ 
    \If{$T_j \neq T_{\text{current}}$} \Comment{Task switching cost applies}
        \State Increment $C_\text{total} = C_\text{total} + c_\text{switch}(T_{\text{current}}, T_j)$
        \State Update $T_{\text{current}} = T_j$
    \EndIf
\EndFor
\end{algorithmic}
\end{algorithm}

\textbf{Active Experiment Selection}.
% As shown in Algorithm~\ref{alg:short_pseudocode}, we use this acquisition function to iteratively sample experiments.
We use this acquisition function to iteratively sample experiments, as shown in Algorithm~\ref{alg:short_pseudocode}.
To mitigate the cold-start problem in active learning, we initialize the dataset $\mathcal{D}$ with a a single randomly-selected task, for which every policy is evaluated 3 times.
We then train the surrogate model on this data.
At each query step, the acquisition function $a(\pi_i, T_j)$ is computed for all policy-task pairs, which quantifies their informativeness weighted by the cost.
To compute the entropy over model parameters for the EIG metric, we use MC dropout to sample 10 predicted outcome distributions.
To balance exploration and exploitation, we use an epsilon-greedy strategy with a rate of $\epsilon=0.1$.
The selected experiment $(\pi_i, T_j)$ is then executed 3 times, and the observed outcomes are added to the dataset $\mathcal{D}$.
We found in preliminary experiments that 3 trials per selected experiment was often better for cost-efficient population parameter estimation.
Given these new outcomes in the dataset, we keep training the surrogate model on the updated dataset improve its predictions over time.
% In our experiments, we continue this process for a fixed number of queries; however, 

\section{Evaluation on Offline Datasets}

\begin{figure}[!t]
    \centering
    \includegraphics[width=.93\linewidth]{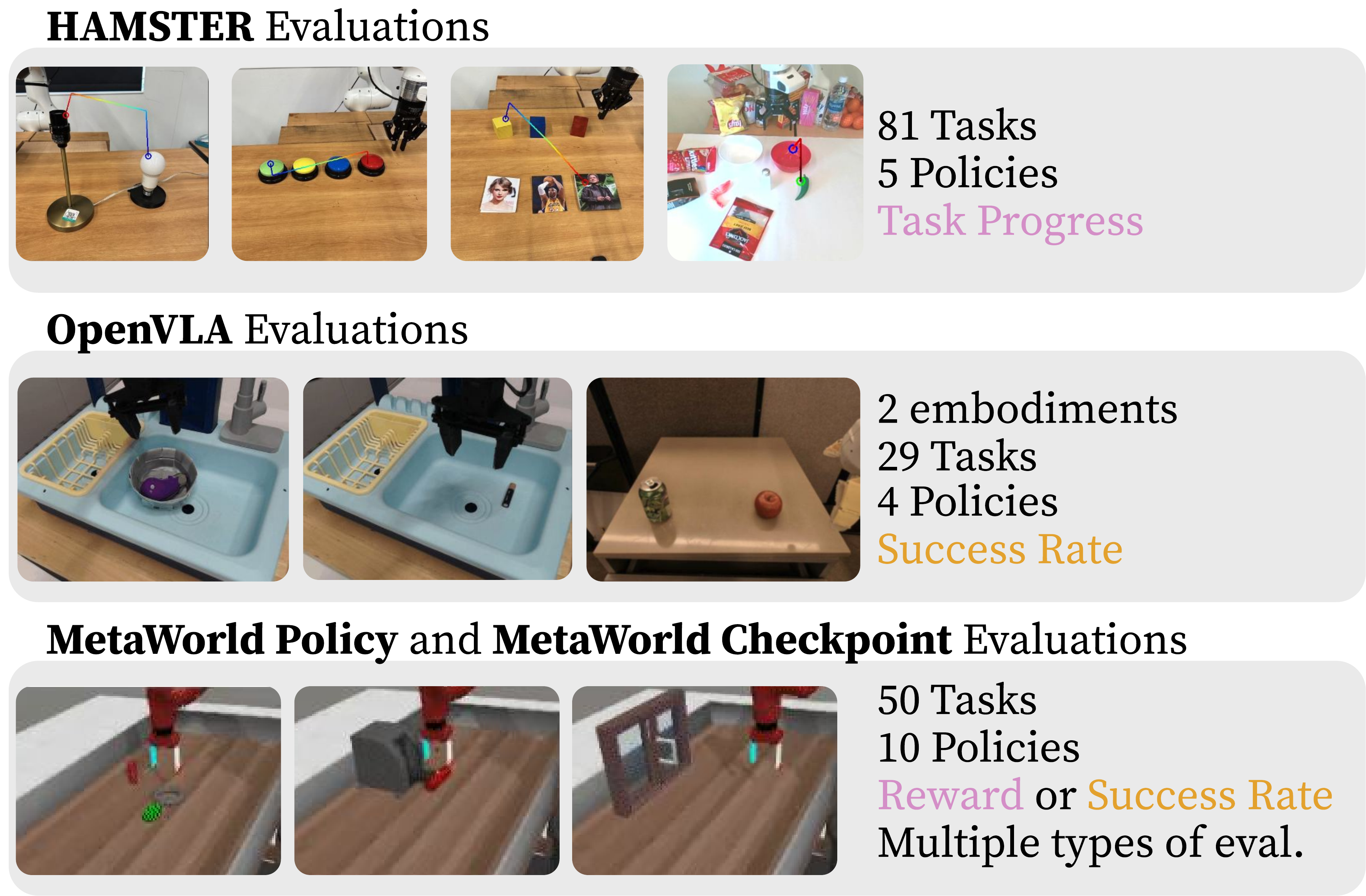}
    \caption{\textbf{Offline Datasets used for Experiments.} We consider 4 settings: (1) evaluations from HAMSTER~\cite{li2025hamster}, (2) evaluations from the OpenVLA paper~\cite{kim24openvla}, (3) MetaWorld~\cite{yu2020meta} where we evaluate different policies, and (4) MetaWorld where we evaluate multiple checkpoints of a single policy. For the MetaWorld evaluations, we can model the performance distributions of success rate or continuous rewards. For OpenVLA, the outcomes are binary success rate. For HAMSTER, evaluations were run over a large number of tasks only once while tracking only task progress, so we use this mean value as a mean for a unimodal Gaussian and a fixed standard deviation.
    }
    \label{fig:datasets}
\end{figure}

To evaluate our active testing framework, we leverage evaluations that have already been conducted which we then sample offline.
We use experiments from the HAMSTER paper~\cite{li2025hamster}, the  OpenVLA paper~\cite{kim24openvla}, and from MetaWorld~\cite{yu2020meta}, as visualized in Figure~\ref{fig:datasets}.
For MetaWorld, we train two versions, one focused on understanding our framework's ability in evaluating different policies and another on evaluating multiple checkpoints of a single policy.
Each of these datasets can be modeled with different underlying distributions and have varying costs, semantic diversity, and skills.
More details on training for MetaWorld, switching costs for the datasets, and other details can be found in Appendix~\ref{app:offline_datasets}.

\textbf{HAMSTER.}
We use evaluations from the HAMSTER paper~\cite{li2025hamster}, which evaluates a hierarchical VLA model against 4 other policies such as OpenVLA~\cite{kim24openvla} and Octo~\cite{octo_2023} across 81 tasks. 
These 81 tasks are of varying complexity, with diverse task types, objects, and linguistic variation that were evaluated once each.
Their work uses a continuous task progress metric; however, since they only evaluated each policy-task pair once, we treat the single continuous value as the mean of a Gaussian distribution with a fixed standard deviation.
For switching cost, we add an additional cost if the policy switches from one task type to another. 
More details on this cost can be found in Appendix~\ref{app:offline_datasets}.
% The success of that one policy-task pair evaluation is treated as the mean success metric for that experiment. 

\textbf{OpenVLA.}
We use evaluations from the OpenVLA paper~\cite{kim24openvla}, which compares 4 policies over 29 tasks. 
In their paper, some tasks allow for partial success (0.5). 
For simplicity, we round the partial successes down to maintain a binary success metric.
OpenVLA also provides results across two embodiments.
Therefore, in addition to a higher cost term to switching tasks that require a large scene reset, we add an additional cost term to switch between embodiments. More details in Appendix~\ref{app:offline_datasets}.

Given these datasets, we show that the types of policy and task representations that are useful for active learning, and then we can leverage the surrogate model for cost-aware active experiment selection. 
\begin{figure*}[t]
    \centering
    % placeholder
    \includegraphics[width=1\linewidth]{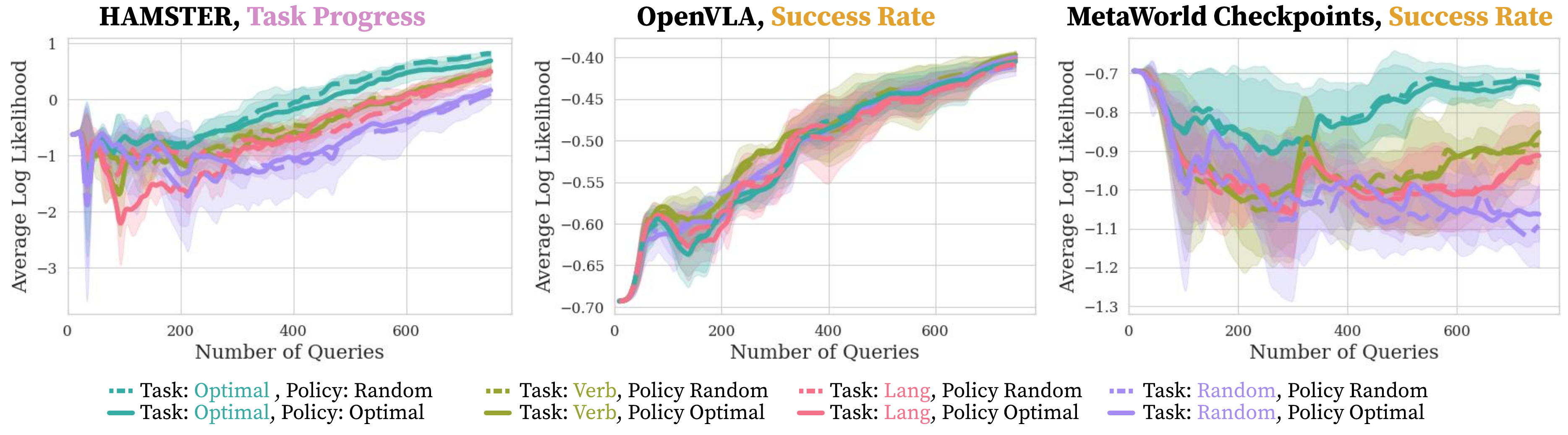}
    \caption{\textbf{Task and Policy Representation Experiments.} 
    We compute the average log likelihood of all outcomes under probability distribution represented by the predicted population parameters across various policy and task representations. We evaluate these methods over the HAMSTER, OpenVLA, and MetaWorld Checkpoints offline evaluation datasets over \textcolor{mypink}{continuous} and \textcolor{mygold}{binary} performance distributions.
    We find no large difference between random or optimal embeddings as a policy representation, indicating that there is not much shared information between policies.
    However, we find that for task representation, \textcolor{optimal}{\textbf{Optimal}} consistently perform the best, followed by \textcolor{verb}{\textbf{Verb}}, then \textcolor{lang}{\textbf{Lang}}, and lastly \textcolor{random}{\textbf{Random}}.
    Language-based embeddings is a good task representation that we can leverage for better active learning.
    % We find that optimal embeddings for tasks and policies perform the best, while random embeddings perform the worst. 
    % We find that language embeddings perform better as a task representation than random embeddings as it gets closer to the rate at which optimal embeddings improve.
    % \vspace{-1em}
    }
    \label{fig:model_exps}
\end{figure*}

\textbf{MetaWorld Policies.}
MetaWorld~\cite{yu2020meta} is an open-source simulated benchmark containing a set of 50 different manipulation environments for multi-task learning.
We train 10 policies on every environment with different policy architecture sizes and varying amounts of noise in the robot's state to create robot policies with diverse behaviors.
We then collected 100 trajectories of each policy-task pair to serve as an approximation of the true performance population distribution.
By using the MetaWorld simulator, we can estimate performance distributions for binary success rate and a continuous reward normalized between 0 and 1.
The switching cost is set based on whether the target object of the scene, such as a drawer, is swapped out for another object, like a lever.
This dataset allows us to understand how our framework can learn the performance distributions across diverse policies.

\textbf{MetaWorld Checkpoints.}
Evaluation on a robot is not only used for comparing policies, but also to find the best checkpoints.
As such, we train a single state-based MetaWorld policy, store 11 checkpoints over the training process, and then evaluate them.
In preliminary experiments, we found that the checkpoint-based setting has a lower-rank structure in terms of the performance distributions.
This offline dataset allows us to exploit the shared information across policies.

Given these datasets, we will discuss two experiments in the next two sections: that shows that language is an informative prior in modeling the performance relationships between tasks, and that our surrogate model can be used for cost-aware experiment selection.

% . Given an evaluation score of 7.5, for example, we generate a sequence of seven successes (1s) and three failures (0s).

% This trajectory data was made of pairs of a state and the action taken from that state. Then, for each environment, we generated a natural language query that describes the task in the environment. We then trained an MLP-based model using the natural language query, embedded as a 768-dimensional vector, and the 39-dimensional state vector in the trajectory to output the 4-dimensional action vector. The resulting model is a language-conditioned behavior-cloned policy trained on all the language instructions and environment rollouts. This model is effectively trained to be a single generalist language-conditioned policy for the Metaworld environment.

% We trained 10 of these generalist policies for Metaworld, varying the training data with randomness to ensure each policy was fairly distinct from another, and then rolled out each of them 100 times each on each of the 50 multi-task learning environments. The success and reward data is used in our parameter estimations for policy evaluation. 

% \input{text/4_method}

\section{Task and Policy Representation}
\label{sec:task_results}
As we define an experiment based on a task and a policy, we must design different embedding strategies for each of them.
We first discuss baselines and upper bounds on task and policy representations, then we show results on how these representations impact our surrogate model.

\subsection{Experimental Setup}
As it is unclear what an ideal representation for a policy or task is, we compute an upper bound for a task and policy representation by taking all the pre-evaluated outcomes, and then training learnable embeddings on the task of estimating performance. 
Thus, these task and policy representations have specifically been tuned for this prediction task.
We can then use these learned embeddings as optimal representations of the task and policy.

However, this optimal approach requires all the data a priori. 
Thus, we need a way to represent both a task and a policy.
The most direct way to represent a task is based on the language description of a task.
As described in Section~\ref{sec:method}, we define our task representation as a weighted sum between the language embeddings of the task description and the verbs.
We call this approach \textcolor{verb}{\textbf{Verb}}.
Overall, we consider the following task representation types as upper bounds and baselines:
\begin{enumerate}
    \item \textcolor{optimal}{\textbf{Optimal:}} Leverage all the data a priori to learn embeddings that are useful for predicting performance;
    \item \textcolor{verb}{\textbf{Verb:}} Use a weighted sum of the language embedding of the task and the language embedding of its verbs;
    \item \textcolor{lang}{\textbf{Language:}} Use a language embedding of the task as its representation; and
    \item \textcolor{random}{\textbf{Random:}} Assume no relationship between policies and tasks by using random embeddings.
\end{enumerate}

\begin{figure*}[!ht]
    \centering
    \includegraphics[width=.95\linewidth]{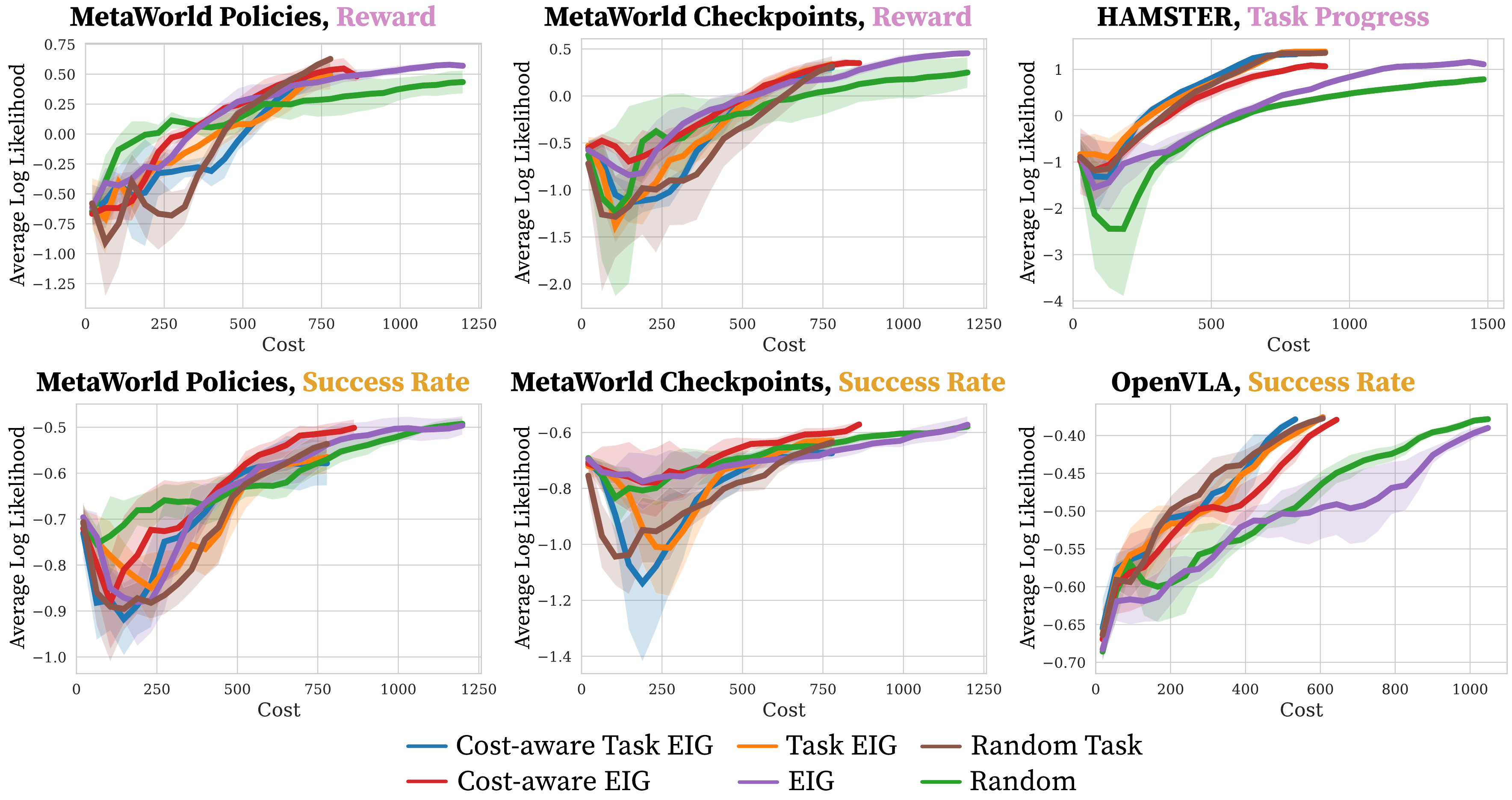}
    \caption{\textbf{Average Log Likelihood Over Cost.} 
    We show the average log likelihood of all the outcomes in our offline dataset against the cost of evaluation for MetaWorld Policies, MetaWorld Checkpoints, HAMSTER, and OpenVLA over \textcolor{mypink}{continuous} and \textcolor{mygold}{binary} performance distributions.
    Each set of experiments is run for 1500 trials.
    % We find that EIG-based approaches generally outperform random baselines, both in the task- and policy-task-based sampling.
    We find that EIG-based approaches struggle to model the true distribution in a more cost-efficient manner than Random Task sampling.
    Task-based sampling strategies are more cost-efficient than policy-task approaches.
    % \vspace{-1em}
    }
    \label{fig:cost_exps}
\end{figure*}

Unlike a task representation through language, there is no clear representation for a policy. 
We leave the exploration of new policy representations to future work and focus on two policy representations: \textbf{Optimal} and \textbf{Random}.

All experiments in this section were run for 750 evaluation steps over three seeds.
To evaluate how much these embeddings improve the performance of population parameter estimation during active experiment selection, we look at the log likelihood of all the outcomes in our offline dataset against a probability distribution represented by the predicted population parameters from the surrogate model.
Each experiment is sampled similar to how researchers typically evaluate: we select a random task and test each policy three times.
% \lipsum[2]

\subsection{Results}

We evaluate the effectiveness of different task representations by computing the average log likelihood of the full dataset against the predicted distribution across multiple datasets, including MetaWorld Policies, MetaWorld Checkpoints, OpenVLA, and HAMSTER, as shown in Figure \ref{fig:model_exps}.

\textbf{Random representations do not share information across policies and tasks.}
Our results indicate that random embeddings consistently perform worse, as they fail to capture any meaningful structure or shared information between tasks. 
In contrast, optimal embeddings, which used the entire dataset to tune its representation, outperforms all baselines. 
We found that the increasing performance of random performance is due to new experiments being sampled; however, minimal interpolation of outcomes occurred.

\textbf{Task representations vary depending on the kinds of tasks.}
We find that the types of tasks matter. 
The HAMSTER evaluations consist of many changes to objects rather than changes to the type of task itself such as ``pickup the milk\dots" and ``pickup the shrimp\dots"
This structure leads to clearer benefits when using language-based representations.
In contrast, OpenVLA has less separable tasks, thus it shows a much smaller separation between random, optimal, and language-based embeddings.
Metaworld Checkpoints, however, show a more stable improvement of \textbf{Verb} as opposed to simply \textbf{Lang} since there are many more tasks.
% A ``pickup orange mug" task and a ``push orange mug" task would have similar performance even if they are different verbs. 
% We find that the verb-specific information in the \textbf{Verb} representation is a better

\textbf{Language does not explain all the shared information between tasks.}
Despite the improvement from using language or verbs as a task representation, they do not fully bridge the gap to optimal embeddings.
The difference between the optimal embeddings and language embeddings indicates that task descriptions, even when focused on the verbs, do not capture all the information to describe a task's relationship to its performance.
Our approach does not include the observations of the trajectory, and this difference between optimal and language embeddings may be explained by the lack of the initial image.
We leave it to future work to explore this direction.
% We believe it would be interesting future work to  remaining information is likely captured in the observation itself, which our framework does not include.

\textbf{Optimal policy embeddings do not provide meaningful gains.}
While task embeddings provide a meaningful way to represent tasks, we found that random or optimal policy embeddings do not provide any significant improvements compared to one another.
This result may be due to the procedure for learning the optimal embeddings overly relying on the task embeddings during their training, or may be caused by the relatively small number of policies that were evaluated, which ranged from 4 to 11. 
In contrast, there were between 29 to 81 tasks that were evaluated against, so there was higher overlap between some tasks.

\begin{figure*}[t]
    \centering
    \includegraphics[width=.95\linewidth]{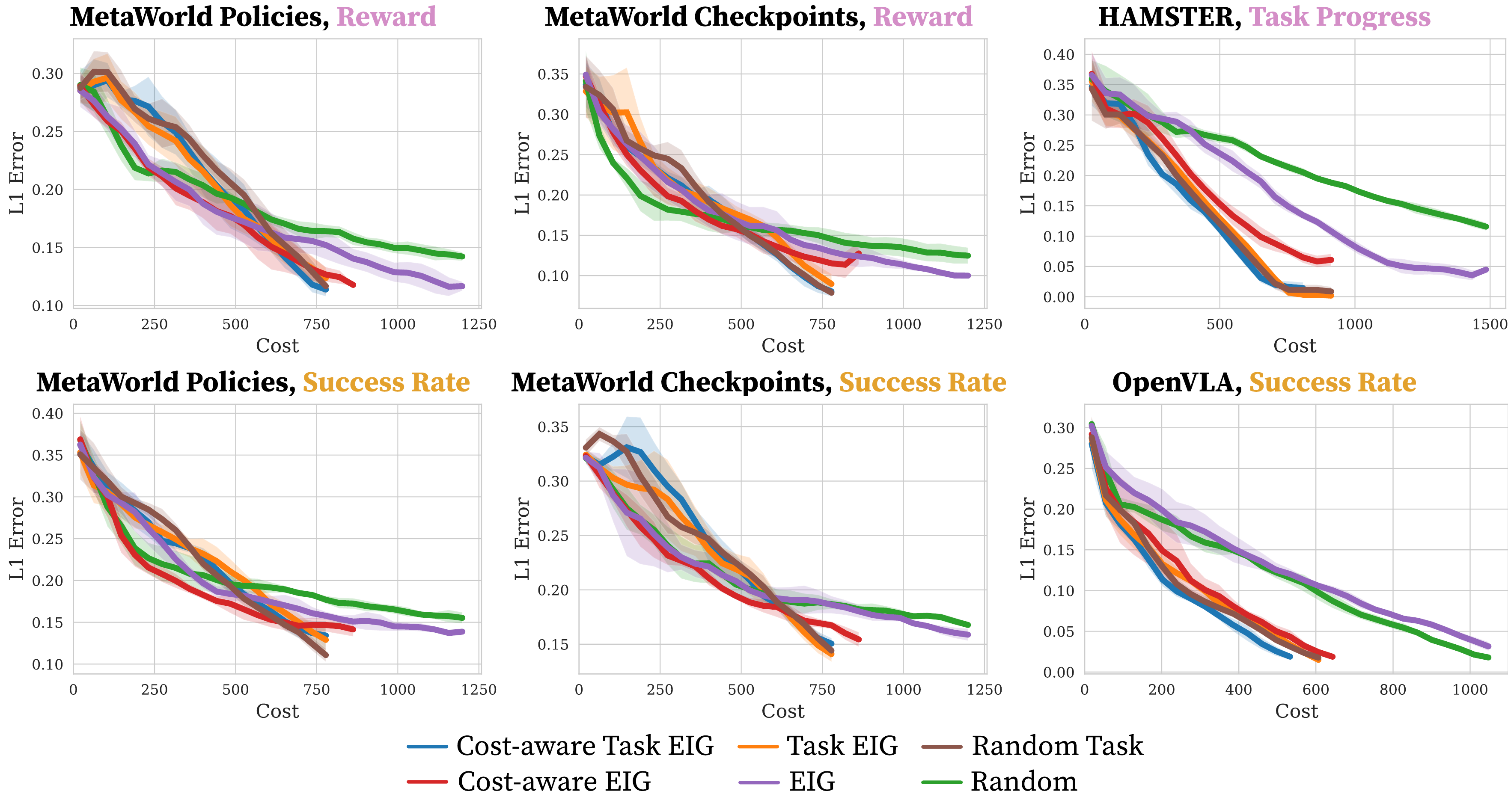}
    \caption{\textbf{Average L1 Error of the Mean Over Cost.} Instead of computing the average log likelihood of the data as in Figure~\ref{fig:cost_exps}, we compute the error between the ground truth means of a policy-task pair and the mean of the predicted probability distribution. 
    In this case, we find that our method is better able to estimate the means for both \textcolor{mypink}{continuous} and \textcolor{mygold}{binary} distributions.
    We find that task sampling methods are generally more cost-efficient for the same average log likelihood than the policy-task sampling methods.
    % \vspace{-1em}
    }
    \label{fig:l1_exps}
\end{figure*}

\section{Cost-Aware Experiment Selection}

To evaluate the effectiveness of our cost-aware active experiment selection methods, we assess the population parameter estimation capability of our framework across various datasets using continuous and binary performance distributions.

\subsection{Experimental Setup}

\textbf{Sampling Strategies.}
To select the most informative experiment based on an acquisition function $a(\pi_i, T_j)$, we must design acquisition functions to define our sampling strategy.
We consider two types of sampling strategies. 
The first is to select both a policy and a task to run an evaluation on. 
Given the EIG formulation in Section~\ref{sec:method}, we define three sampling strategies with this approach:
\begin{itemize}
    \item \textcolor{rand}{\textbf{Random Sampling:}} Select a task-policy pair uniformly at random $a(\pi_i, T_j)=1/ (|\mathcal{P}| \times |\mathcal{T}|)$;
    \item  \textcolor{eig}{\textbf{EIG:}} Select a task-policy pair $(\pi_i,t_j)$ with the highest EIG: $a(\pi_i, T_j)=\mathcal{I}(\pi_i,T_j)$;
    \item  \textcolor{cost_eig}{\textbf{Cost-aware EIG:}} Select a task-policy pair that maximizes the cost-aware EIG according to Equation~\ref{eq:cost_aware_eig}.
\end{itemize}

The second type of sampling strategy is to select a task, and then evaluate every policy in that task $d=3$ times.

\begin{itemize}
    \item  \textcolor{rand_task}{\textbf{Random Task:}} Select a task uniformly at random and evaluate all policies on that task: $a(t_j) = 1 / |\mathcal{T}|$
    \item  \textcolor{task_eig}{\textbf{Task EIG:}} Select a task $T_j$ that maximizes the summed EIG across all policies: $a(t_j)=\sum_{i} \mathcal{I}(\pi_i,T_j)$
    \item  \textcolor{cost_task_eig}{\textbf{Cost-aware Task EIG:}} Select a task $T_j$ that maximizes the summed cost-aware EIG across all policies: $a(T_j)=\sum_{i} a_{\text{cost-aware}}(\pi_i,T_j, T_\text{current})$
\end{itemize}

The task-based sampling strategies is more realistic to how experimenters evaluate their robots today, as experimenters typically select a task and then evaluate every policy.
% However, it may be optimal to evaluate policy-task sampling as it may require less execution cost.

We evaluated each method for 1500 evaluation steps over three seeds using \textbf{Random} policy embeddings and \textbf{Verb} task embeddings.
To evaluate these methods, we consider two metrics: (1) the log likelihood of all the outcomes in our offline dataset against the predicted population parameters of the model, and (2) the L1 error between the mean from all the data for a policy-task pair against the mean derived from the estimated population parameters.

\subsection{Results}

% Describe the importance of language, etc. Discuss hypothesis.
% \lipsum[2]
% \lipsum[2]

% \subsection{Cost over time}

% Discuss results of Figure \ref{fig:cost_exps} and Figure~\ref{fig:l1_exps}.
\begin{figure*}[t]
    \centering
    \includegraphics[width=.94\linewidth]{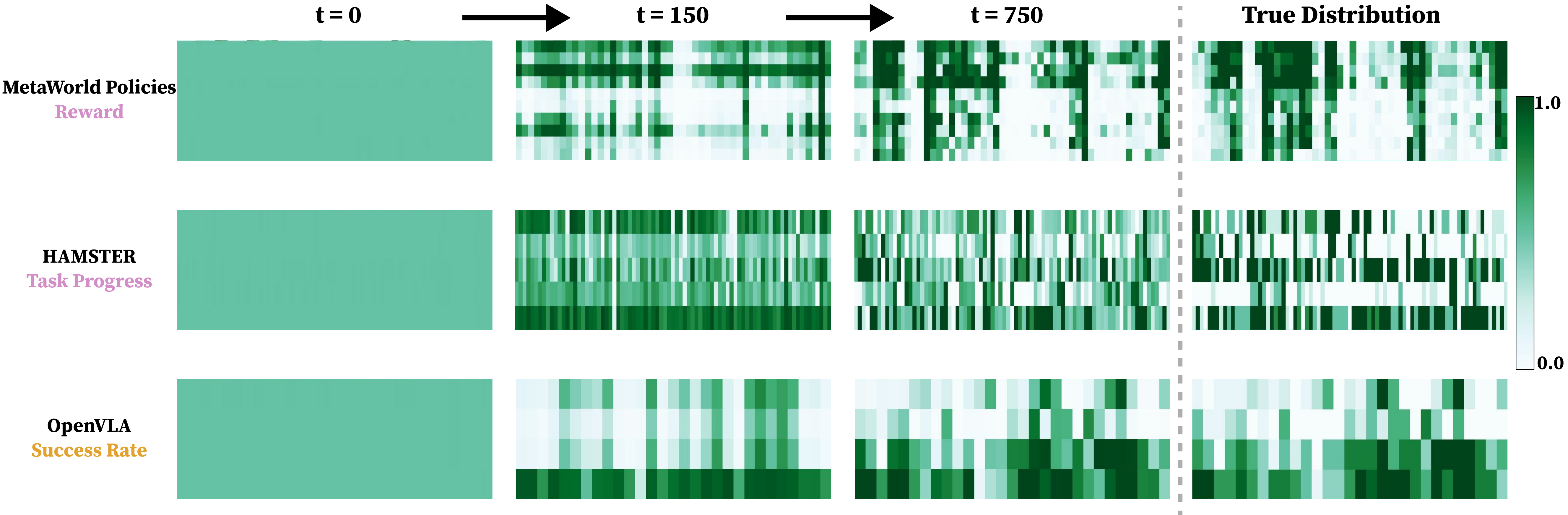}
    \caption{\textbf{Predicted Mean Distributions.}
    We provide a visualization of the means for the predicted \textcolor{mypink}{continuous} and \textcolor{mygold}{binary} distributions over 0, 150, and 750 sampled queries. 
    We use random sampling with 3 evaluations per policy-task pair to show that our surrogate model can actively learn the full distribution of performance as well as have a good understanding of the performance distribution over time.
    For example, for MetaWorld Policies at $t=750$, $750/3=250$ policy-task pairs were sampled of the total $50*10=500$ possible policy-task pairs that could be evaluated, the estimated mean performance is qualitatively comparable to the true mean; Figure~\ref{fig:l1_exps} reports these results quantitatively as L1 error.
    % the estimation of the mean performance looks comparable to the true mean.
    }
    \label{fig:pred_dists}
\end{figure*}

\textbf{EIG-based approaches struggle to learn population parameters that represent all the data, but better estimate the mean.}
In Figure~\ref{fig:cost_exps}, we show the average log likelihood of all the outcomes in our offline dataset against the probability distribution represented by the predicted population parameters from the surrogate model.
In both task- and policy-task sampling approaches, we find that EIG-based approaches fit the original data marginally better than random baselines.
In some cases, such as for MetaWorld Policies with success rate, cost-aware EIG is able to maintain a larger improvement; however, this result is not consistent across other datasets.
This result indicates that learning this full underlying distribution remains challenging, particularly in the early stages of evaluation when data is sparse. 
However, in Figure~\ref{fig:l1_exps}, EIG-based approaches clearly dominate when estimating the mean of these distributions, and often are able to estimate the mean at a lower cost compared to random baselines.
If the cost is fixed at a lower value, as if it was a maximum cost-budget, then we find that EIG-based approaches better estimate the means.
% We find that at costs that are earlier, EIG-based approaches better estimate the means at a lower cost.
% Even if the full distribution is difficult to capture, the mean estimates are useful.

\textbf{Tradeoffs between task- and policy-task sampling.}
Both Figure~\ref{fig:cost_exps} and Figure~\ref{fig:l1_exps} show that task-based sampling is generally better in OpenVLA and HAMSTER, but cost-aware EIG is generally estimates the L1 error better than its task-based counterpart on MetaWorld.
Policy-task sampling approaches are likely more efficient in MetaWorld experiments as there are a large number of experiments where there is a high cost to switch, and evaluating 10 policies over a single task may not be as informative.
In contrast, HAMSTER and OpenVLA have fewer policies, meaning the cost of evaluating all policies for a single task is lower.
% Since we assume a small uniform cost to evaluating a policy within the same task, there is a negligible penalty to task-based sampling methods.
Additionally, we found that policy-task sampling methods are more likely to switch tasks, causing a faster accumulation of cost.

\textbf{Learning the Performance Landscape.}
Figure~\ref{fig:pred_dists} illustrates how our formulation of sequentially sampling experiments progressively refines the predictions of the performance landscape. 
Early in the evaluation process, predictions are generally around the mean and are misaligned with the true distribution.
As more experiments are selected, the means begin to resemble the true mean distribution.

% \subsection{Noise Robustness}
% Discuss noisy gaussian setting \ref{fig:noisy_gaussian}
% \lipsum[1]

% \textbf{Noisiness affects stuff somehow}
% \lipsum[2]

% \input{figs_tex/noisy_gaussian}

% \section{Practical Advice on Using Active Testing}
% \abrar{write this!}
% \textbf{How you do modelling maters a lot}
% \lipsum[2]

% \textbf{Designing your text encoder matters a lot}
% \lipsum[2]

\section{Conclusion and Limitations}

We present a framework for the efficient evaluation of multitask robot policies. 
By framing evaluation as an active testing problem, we develop techniques that use relationships between tasks to predict policy performance distributions. 
In particular, we focus on methods that select experiments based on the expected information gain.
Our experiments demonstrate that task similarities can indeed be used to predict policy performance in an efficient manner, compared to standard evaluation approaches.
As evaluation settings and policy comparisons continue to scale in size, our methods for active testing can help lower the cost of effective evaluation without sacrificing too much information about policy performance.

\textbf{Future Work.}
To properly be cost-aware, a single look-ahead step is often not enough, as it may be beneficial to plan future evaluations with respect to cost and potential information gain. 
Future work can extend our methods by developing look-ahead algorithms that can select longer sequences of experiments at a time.
In addition, other types of acquisition functions, such as those that batch experiments at once, can be explored.
We focused on ensuring that our surrogate model is able to estimate the landscape of performance across tasks and policies, but future work can focus on other types of comparison, such as finding the best average policy, finding a ranked ordering of policies, or finding the worst performing tasks. 
Each of these would require different active sampling strategies.
Additionally, learning policy embeddings may better predict performance, and policy embedding priors might be formed by encoding the training data of those policies, analogous to ``task embeddings'' in multi-task learning. 
There are also hierarchical relationships between tasks such as ``pour milk" likely depending on being able to ``pick up the milk" that would be exciting to explore in future work.

\textbf{Limitations.}
Though our approach to mitigating the cold-start problem with test-time dropout appears to have improved performance during sampling, this approach has not been rigorously tested by the Bayesian optimization community.
We had also tried other approaches, such as ensembling and variational prediction, but these approaches also overfit to the small size of the dataset early in the evaluation procedure.
We also represented execution costs naively at a fixed cost; however, different tasks may have different execution costs that may depend on whether a policy fails on its task or not, such as having to clean up spilled milk.
Additionally, we chose to use a simple MLP to learn our surrogate model; however, other work often used Bayesian neural networks and Gaussian processes. 
We made this decision because these alternative approaches typically do not scale to larger inputs; however, we did not consider the state-of-the-art for those approaches.
% Though we were skeptical about the scalability of these approaches to language embeddings, we believe such results would have more significant statistical importance.

\section{Acknowledgments}
This work was supported in part by a grant from the Army Research Lab (ARL) Army AI Innovations Institute (A2I2), award number W911NF-23-2-0010.
The claims and findings of this work do not necessarily represent the views of the ARL.

% Qualitatively we saw ensembling overfit to the small amounts of data. Cite some cold start active learning papers.

% Why L1 distance 

% \lipsum[2]

% Cost-Aware Multi-Task Robot Policy Evaluation
% Cost-Aware Evaluation of Robot Policies
% Cost aware robot policy evaluation enables efficient learning of where policies suceced/fail
% Efficiently Estimating Multi-Task Multi-Policy Robot Performance
% Efficient Multi-Task Performance Evaluation

% create a big channel

% \section*{Acknowledgments}

%% Use plainnat to work nicely with natbib. 

\bibliographystyle{plainnat}
\bibliography{references}

\begin{thebibliography}{52}
\providecommand{\natexlab}[1]{#1}
\providecommand{\url}[1]{\texttt{#1}}
\expandafter\ifx\csname urlstyle\endcsname\relax
  \providecommand{\doi}[1]{doi: #1}\else
  \providecommand{\doi}{doi: \begingroup \urlstyle{rm}\Url}\fi

\bibitem[Anderson et~al.(2021)Anderson, Shrivastava, Truong, Majumdar, Parikh, Batra, and Lee]{anderson2021sim}
Peter Anderson, Ayush Shrivastava, Joanne Truong, Arjun Majumdar, Devi Parikh, Dhruv Batra, and Stefan Lee.
\newblock Sim-to-real transfer for vision-and-language navigation.
\newblock \emph{Conference on Robot Learning (CoRL)}, 2021.

\bibitem[Anwar et~al.(2024)Anwar, Gupta, and Thomason]{anwar2024contrast}
Abrar Anwar, Rohan Gupta, and Jesse Thomason.
\newblock Contrast sets for evaluating language-guided robot policies.
\newblock \emph{Conference on Robot Learning (CoRL)}, 2024.

\bibitem[Anwar et~al.(2025)Anwar, Welsh, Biswas, Pouya, and Chang]{anwar2024remembr}
Abrar Anwar, John Welsh, Joydeep Biswas, Soha Pouya, and Yan Chang.
\newblock Remembr: Building and reasoning over long-horizon spatio-temporal memory for robot navigation.
\newblock \emph{International Conference on Robotics and Automation (ICRA)}, 2025.

\bibitem[Bhatt et~al.(2023)Bhatt, Nemlekar, Fontaine, Tjanaka, Zhang, Hsu, and Nikolaidis]{bhatt2023surrogate}
Varun Bhatt, Heramb Nemlekar, Matthew~C Fontaine, Bryon Tjanaka, Hejia Zhang, Ya-Chuan Hsu, and Stefanos Nikolaidis.
\newblock Surrogate assisted generation of human-robot interaction scenarios.
\newblock \emph{Conference on Robot Learning (CoRL)}, 2023.

\bibitem[Bishop(1994)]{bishop1994mixture}
Christopher~M Bishop.
\newblock Mixture density networks.
\newblock \emph{Technical Report}, 1994.

\bibitem[Black et~al.(2024)Black, Brown, Driess, Esmail, Equi, Finn, Fusai, Groom, Hausman, Ichter, et~al.]{black2024pi_0}
Kevin Black, Noah Brown, Danny Driess, Adnan Esmail, Michael Equi, Chelsea Finn, Niccolo Fusai, Lachy Groom, Karol Hausman, Brian Ichter, et~al.
\newblock $\pi_0 $: A vision-language-action flow model for general robot control.
\newblock \emph{arXiv preprint arXiv:2410.24164}, 2024.

\bibitem[Brochu et~al.(2010)Brochu, Cora, and De~Freitas]{brochu2010tutorial}
Eric Brochu, Vlad~M Cora, and Nando De~Freitas.
\newblock A tutorial on bayesian optimization of expensive cost functions, with application to active user modeling and hierarchical reinforcement learning.
\newblock \emph{arXiv preprint arXiv:1012.2599}, 2010.

\bibitem[Chang et~al.(2024)Chang, Wang, Wang, Wu, Yang, Zhu, Chen, Yi, Wang, Wang, et~al.]{chang2024survey}
Yupeng Chang, Xu~Wang, Jindong Wang, Yuan Wu, Linyi Yang, Kaijie Zhu, Hao Chen, Xiaoyuan Yi, Cunxiang Wang, Yidong Wang, et~al.
\newblock A survey on evaluation of large language models.
\newblock \emph{ACM Transactions on Intelligent Systems and Technology (TIST)}, 2024.

\bibitem[Cozad et~al.(2014)Cozad, Sahinidis, and Miller]{cozad2014learning}
Alison Cozad, Nikolaos~V Sahinidis, and David~C Miller.
\newblock Learning surrogate models for simulation-based optimization.
\newblock \emph{AIChE Journal}, 60\penalty0 (6):\penalty0 2211--2227, 2014.

\bibitem[Deitke et~al.(2020)Deitke, Han, Herrasti, Kembhavi, Kolve, Mottaghi, Salvador, Schwenk, VanderBilt, Wallingford, Weihs, Yatskar, and Farhadi]{deitke2020robothor}
Matt Deitke, Winson Han, Alvaro Herrasti, Aniruddha Kembhavi, Eric Kolve, Roozbeh Mottaghi, Jordi Salvador, Dustin Schwenk, Eli VanderBilt, Matthew Wallingford, Luca Weihs, Mark Yatskar, and Ali Farhadi.
\newblock {RoboTHOR: An Open Simulation-to-Real Embodied AI Platform}.
\newblock \emph{{Conference on Computer Vision and Pattern Recognition (CVPR)}}, 2020.

\bibitem[Eggensperger et~al.(2015)Eggensperger, Hutter, Hoos, and Leyton-Brown]{eggensperger2015efficient}
Katharina Eggensperger, Frank Hutter, Holger Hoos, and Kevin Leyton-Brown.
\newblock Efficient benchmarking of hyperparameter optimizers via surrogates.
\newblock In \emph{Proceedings of the AAAI conference on artificial intelligence}, 2015.

\bibitem[Gal and Ghahramani(2016)]{gal2016dropout}
Yarin Gal and Zoubin Ghahramani.
\newblock Dropout as a bayesian approximation: Representing model uncertainty in deep learning.
\newblock \emph{International Conference on Machine Learning (ICML)}, 2016.

\bibitem[Gao et~al.(2024)Gao, Xie, Xiao, Finn, and Sadigh]{gao2024}
Jensen Gao, Annie Xie, Ted Xiao, Chelsea Finn, and Dorsa Sadigh.
\newblock {Efficient Data Collection for Robotic Manipulation via Compositional Generalization}.
\newblock \emph{Proceedings of Robotics: Science and Systems (RSS)}, 2024.

\bibitem[Gardner et~al.(2020)Gardner, Artzi, Basmov, Berant, Bogin, Chen, Dasigi, Dua, Elazar, Gottumukkala, Gupta, Hajishirzi, Ilharco, Khashabi, Lin, Liu, Liu, Mulcaire, Ning, Singh, Smith, Subramanian, Tsarfaty, Wallace, Zhang, and Zhou]{gardner2020evaluating}
Matt Gardner, Yoav Artzi, Victoria Basmov, Jonathan Berant, Ben Bogin, Sihao Chen, Pradeep Dasigi, Dheeru Dua, Yanai Elazar, Ananth Gottumukkala, Nitish Gupta, Hannaneh Hajishirzi, Gabriel Ilharco, Daniel Khashabi, Kevin Lin, Jiangming Liu, Nelson~F. Liu, Phoebe Mulcaire, Qiang Ning, Sameer Singh, Noah~A. Smith, Sanjay Subramanian, Reut Tsarfaty, Eric Wallace, Ally Zhang, and Ben Zhou.
\newblock Evaluating models{'} local decision boundaries via contrast sets.
\newblock \emph{Findings of Empirical Methods in Natural Language Processing (EMNLP Findings)}, 2020.

\bibitem[Gervet et~al.(2023)Gervet, Chintala, Batra, Malik, and Chaplot]{gervet2023navigating}
Theophile Gervet, Soumith Chintala, Dhruv Batra, Jitendra Malik, and Devendra~Singh Chaplot.
\newblock Navigating to objects in the real world.
\newblock \emph{Science Robotics}, 2023.

\bibitem[Gu et~al.(2024)Gu, Dong, Wei, and Huang]{gu2024minillm}
Yuxian Gu, Li~Dong, Furu Wei, and Minlie Huang.
\newblock Minillm: Knowledge distillation of large language models.
\newblock In \emph{International Conference on Learning Representations (ICLR)}, 2024.

\bibitem[Ha and Schmidhuber(2018)]{ha2018world}
David Ha and J{\"u}rgen Schmidhuber.
\newblock World models.
\newblock \emph{Conference on Neural Information Processing System (NeurIPS)}, 2018.

\bibitem[Hendrycks and Dietterich(2019)]{hendrycks2019benchmarking}
Dan Hendrycks and Thomas Dietterich.
\newblock Benchmarking neural network robustness to common corruptions and perturbations.
\newblock \emph{International Conference on Learning Representations (ICLR)}, 2019.

\bibitem[Hendrycks et~al.(2020)Hendrycks, Liu, Wallace, Dziedzic, Krishnan, and Song]{hendrycks2020pretrained}
Dan Hendrycks, Xiaoyuan Liu, Eric Wallace, Adam Dziedzic, Rishabh Krishnan, and Dawn Song.
\newblock Pretrained transformers improve out-of-distribution robustness.
\newblock \emph{Association for Computational Linguistics (ACL)}, 2020.

\bibitem[Houlsby et~al.(2011)Houlsby, Husz{\'a}r, Ghahramani, and Lengyel]{houlsby2011bayesian}
Neil Houlsby, Ferenc Husz{\'a}r, Zoubin Ghahramani, and M{\'a}t{\'e} Lengyel.
\newblock Bayesian active learning for classification and preference learning.
\newblock \emph{arXiv preprint arXiv:1112.5745}, 2011.

\bibitem[Hu et~al.(2024)Hu, Lucchetti, Schlesinger, Saxena, Freeman, Modak, Guha, and Biswas]{hu2024deploying}
Zichao Hu, Francesca Lucchetti, Claire Schlesinger, Yash Saxena, Anders Freeman, Sadanand Modak, Arjun Guha, and Joydeep Biswas.
\newblock {Deploying and Evaluating LLMs to Program Service Mobile Robots}.
\newblock \emph{IEEE Robotics and Automation Letters (RA-L)}, 2024.

\bibitem[Kadian et~al.(2020)Kadian, Truong, Gokaslan, Clegg, Wijmans, Lee, Savva, Chernova, and Batra]{kadian2020sim2real}
Abhishek Kadian, Joanne Truong, Aaron Gokaslan, Alexander Clegg, Erik Wijmans, Stefan Lee, Manolis Savva, Sonia Chernova, and Dhruv Batra.
\newblock {Sim2Real Predictivity: Does Evaluation in Simulation Predict Real-World Performance?}
\newblock \emph{IEEE Robotics and Automation Letters (RA-L)}, 2020.

\bibitem[Kim et~al.(2024)Kim, Pertsch, Karamcheti, Xiao, Balakrishna, Nair, Rafailov, Foster, Lam, Sanketi, Vuong, Kollar, Burchfiel, Tedrake, Sadigh, Levine, Liang, and Finn]{kim24openvla}
{Moo Jin} Kim, Karl Pertsch, Siddharth Karamcheti, Ted Xiao, Ashwin Balakrishna, Suraj Nair, Rafael Rafailov, Ethan Foster, Grace Lam, Pannag Sanketi, Quan Vuong, Thomas Kollar, Benjamin Burchfiel, Russ Tedrake, Dorsa Sadigh, Sergey Levine, Percy Liang, and Chelsea Finn.
\newblock Openvla: An open-source vision-language-action model.
\newblock \emph{Conference on Robot Learning (CoRL)}, 2024.

\bibitem[Kossen et~al.(2021)Kossen, Farquhar, Gal, and Rainforth]{kossen2021active}
Jannik Kossen, Sebastian Farquhar, Yarin Gal, and Tom Rainforth.
\newblock Active testing: Sample-efficient model evaluation.
\newblock \emph{International Conference on Machine Learning (ICML)}, 2021.

\bibitem[Kress-Gazit et~al.(2024)Kress-Gazit, Hashimoto, Kuppuswamy, Shah, Horgan, Richardson, Feng, and Burchfiel]{kress2024robot}
Hadas Kress-Gazit, Kunimatsu Hashimoto, Naveen Kuppuswamy, Paarth Shah, Phoebe Horgan, Gordon Richardson, Siyuan Feng, and Benjamin Burchfiel.
\newblock Robot learning as an empirical science: Best practices for policy evaluation.
\newblock \emph{arXiv}, 2024.

\bibitem[Ledda et~al.(2023)Ledda, Fumera, and Roli]{ledda2023dropout}
Emanuele Ledda, Giorgio Fumera, and Fabio Roli.
\newblock Dropout injection at test time for post hoc uncertainty quantification in neural networks.
\newblock \emph{Information Sciences}, 2023.

\bibitem[Lee et~al.(2020)Lee, Perrone, Archambeau, and Seeger]{lee2020cost}
Eric~Hans Lee, Valerio Perrone, Cedric Archambeau, and Matthias Seeger.
\newblock Cost-aware bayesian optimization.
\newblock \emph{arXiv preprint arXiv:2003.10870}, 2020.

\bibitem[Li et~al.(2024)Li, Hsu, Gu, Pertsch, Mees, Walke, Fu, Lunawat, Sieh, Kirmani, Levine, Wu, Finn, Su, Vuong, and Xiao]{simpler_env}
Xuanlin Li, Kyle Hsu, Jiayuan Gu, Karl Pertsch, Oier Mees, Homer~Rich Walke, Chuyuan Fu, Ishikaa Lunawat, Isabel Sieh, Sean Kirmani, Sergey Levine, Jiajun Wu, Chelsea Finn, Hao Su, Quan Vuong, and Ted Xiao.
\newblock Evaluating real-world robot manipulation policies in simulation.
\newblock \emph{Conference on Robot Learning (CoRL)}, 2024.

\bibitem[Li et~al.(2025)Li, Deng, Zhang, Jang, Memmel, Garrett, Ramos, Fox, Li, Gupta, and Goyal]{li2025hamster}
Yi~Li, Yuquan Deng, Jesse Zhang, Joel Jang, Marius Memmel, Caelan~Reed Garrett, Fabio Ramos, Dieter Fox, Anqi Li, Abhishek Gupta, and Ankit Goyal.
\newblock Hamster: Hierarchical action models for open-world robot manipulation.
\newblock \emph{International Conference on Learning Representations (ICLR)}, 2025.

\bibitem[Liang et~al.(2022)Liang, Bommasani, Lee, Tsipras, Soylu, Yasunaga, Zhang, Narayanan, Wu, Kumar, et~al.]{liang2022holistic}
Percy Liang, Rishi Bommasani, Tony Lee, Dimitris Tsipras, Dilara Soylu, Michihiro Yasunaga, Yian Zhang, Deepak Narayanan, Yuhuai Wu, Ananya Kumar, et~al.
\newblock Holistic evaluation of language models.
\newblock \emph{Transactions on Machine Learning Research (TLMR)}, 2022.

\bibitem[Loquercio et~al.(2020)Loquercio, Segu, and Scaramuzza]{loquercio2020general}
Antonio Loquercio, Mattia Segu, and Davide Scaramuzza.
\newblock A general framework for uncertainty estimation in deep learning.
\newblock \emph{IEEE Robotics and Automation Letters (RA-L)}, 2020.

\bibitem[Mnih and Salakhutdinov(2007)]{mnih2007probabilistic}
Andriy Mnih and Russ~R Salakhutdinov.
\newblock Probabilistic matrix factorization.
\newblock \emph{Conference on Neural Information Processing Systems (NeurIPS)}, 2007.

\bibitem[Nayyar et~al.(2022)Nayyar, Verma, and Srivastava]{nayyar2022differential}
Rashmeet~Kaur Nayyar, Pulkit Verma, and Siddharth Srivastava.
\newblock Differential assessment of black-box ai agents.
\newblock \emph{AAAI Conference on Artificial Intelligence}, 2022.

\bibitem[{Octo Model Team} et~al.(2024){Octo Model Team}, Ghosh, Walke, Pertsch, Black, Mees, Dasari, Hejna, Xu, Luo, Kreiman, Tan, Chen, Sanketi, Vuong, Xiao, Sadigh, Finn, and Levine]{octo_2023}
{Octo Model Team}, Dibya Ghosh, Homer Walke, Karl Pertsch, Kevin Black, Oier Mees, Sudeep Dasari, Joey Hejna, Charles Xu, Jianlan Luo, Tobias Kreiman, {You Liang} Tan, Lawrence~Yunliang Chen, Pannag Sanketi, Quan Vuong, Ted Xiao, Dorsa Sadigh, Chelsea Finn, and Sergey Levine.
\newblock Octo: An open-source generalist robot policy.
\newblock \emph{Robotics: Science and Systems (RSS)}, 2024.

\bibitem[Parekh et~al.(2024)Parekh, Vitsakis, Suglia, and Konstas]{parekh2024investigating}
Amit Parekh, Nikolas Vitsakis, Alessandro Suglia, and Ioannis Konstas.
\newblock {Investigating the Role of Instruction Variety and Task Difficulty in Robotic Manipulation Tasks}.
\newblock \emph{Conference on Empirical Methods in Natural Language Processing (EMNLP)}, 2024.

\bibitem[Paria et~al.(2020)Paria, Neiswanger, Ghods, Schneider, and P{\'o}czos]{paria2020cost}
Biswajit Paria, Willie Neiswanger, Ramina Ghods, Jeff Schneider, and Barnab{\'a}s P{\'o}czos.
\newblock Cost-aware bayesian optimization via information directed sampling.
\newblock In \emph{Adaptive Experimental Design and Active Learning in the Real World Workshop at ICML}, 2020.

\bibitem[Pumacay et~al.(2024)Pumacay, Singh, Duan, Krishna, Thomason, and Fox]{wilbert_colloseum}
Wilbert Pumacay, Ishika Singh, Jiafei Duan, Ranjay Krishna, Jesse Thomason, and Dieter Fox.
\newblock {THE COLOSSEUM: A Benchmark for Evaluating Generalization for Robotic Manipulation}.
\newblock \emph{Robotics: Science and Systems (RSS)}, 2024.

\bibitem[Qian et~al.(2006)Qian, Seepersad, Joseph, Allen, and Jeff~Wu]{qian2006building}
Zhiguang Qian, Carolyn~Conner Seepersad, V~Roshan Joseph, Janet~K Allen, and CF~Jeff~Wu.
\newblock Building surrogate models based on detailed and approximate simulations.
\newblock \emph{Journal of Mechanical Design}, 2006.

\bibitem[Rainforth et~al.(2024)Rainforth, Foster, Ivanova, and Bickford~Smith]{rainforth2024modern}
Tom Rainforth, Adam Foster, Desi~R Ivanova, and Freddie Bickford~Smith.
\newblock Modern bayesian experimental design.
\newblock \emph{Statistical Science}, 39\penalty0 (1):\penalty0 100--114, 2024.

\bibitem[Recht et~al.(2019)Recht, Roelofs, Schmidt, and Shankar]{recht2019imagenet}
Benjamin Recht, Rebecca Roelofs, Ludwig Schmidt, and Vaishaal Shankar.
\newblock Do imagenet classifiers generalize to imagenet?
\newblock In \emph{International Conference on Machine Learning (ICML)}, 2019.

\bibitem[Sawade et~al.(2010)Sawade, Landwehr, Bickel, and Scheffer]{sawade2010active}
Christoph Sawade, Niels Landwehr, Steffen Bickel, and Tobias Scheffer.
\newblock Active risk estimation.
\newblock In \emph{Proceedings of the 27th International Conference on Machine Learning (ICML-10)}, pages 951--958, 2010.

\bibitem[Shah et~al.(2022)Shah, Sridhar, Dashora, Stachowicz, Black, Hirose, and Levine]{shah2023vint}
Dhruv Shah, Ajay Sridhar, Nitish Dashora, Kyle Stachowicz, Kevin Black, Noriaki Hirose, and Sergey Levine.
\newblock Vint: A foundation model for visual navigation.
\newblock \emph{Conference on Robot Learning (CoRL)}, 2022.

\bibitem[Shah et~al.(2023)Shah, Osi{\'n}ski, Levine, et~al.]{shah2023lm}
Dhruv Shah, B{\l}a{\.z}ej Osi{\'n}ski, Sergey Levine, et~al.
\newblock Lm-nav: Robotic navigation with large pre-trained models of language, vision, and action.
\newblock \emph{Conference on Robot Learning (CoRL)}, 2023.

\bibitem[Shahriari et~al.(2015)Shahriari, Swersky, Wang, Adams, and De~Freitas]{shahriari2015taking}
Bobak Shahriari, Kevin Swersky, Ziyu Wang, Ryan~P Adams, and Nando De~Freitas.
\newblock Taking the human out of the loop: A review of bayesian optimization.
\newblock \emph{Proceedings of the IEEE}, 104\penalty0 (1):\penalty0 148--175, 2015.

\bibitem[Tosh et~al.(2022)Tosh, Tec, and Tansey]{tosh2022targeted}
Christopher Tosh, Mauricio Tec, and Wesley Tansey.
\newblock Targeted active learning for probabilistic models.
\newblock \emph{arXiv preprint arXiv:2210.12122}, 2022.

\bibitem[Verma et~al.(2021)Verma, Marpally, and Srivastava]{verma2021discovering}
Pulkit Verma, Shashank~Rao Marpally, and Siddharth Srivastava.
\newblock Discovering user-interpretable capabilities of black-box planning agents.
\newblock \emph{International Conference on Principles of Knowledge Representation and Reasoning (KR)}, 2021.

\bibitem[Verma et~al.(2023)Verma, Karia, and Srivastava]{verma2023autonomous}
Pulkit Verma, Rushang Karia, and Siddharth Srivastava.
\newblock Autonomous capability assessment of sequential decision-making systems in stochastic settings.
\newblock \emph{Conference on Neural Information Processing Systems (NeurIPS)}, 2023.

\bibitem[Vincent et~al.(2024)Vincent, Nishimura, Itkina, Shah, Schwager, and Kollar]{that_tri_paper}
Joseph~A Vincent, Haruki Nishimura, Masha Itkina, Paarth Shah, Mac Schwager, and Thomas Kollar.
\newblock {How Generalizable Is My Behavior Cloning Policy? A Statistical Approach to Trustworthy Performance Evaluation}.
\newblock \emph{IEEE Robotics and Automation Letters (RA-L)}, 2024.

\bibitem[Wang et~al.(2021)Wang, Ding, Li, and Zheng]{wang2021cline}
Dong Wang, Ning Ding, Piji Li, and Hai-Tao Zheng.
\newblock {CLINE: Contrastive Learning with Semantic Negative Examples for Natural Language Understanding}.
\newblock \emph{Association for Computational Linguistics (ACL)}, 2021.

\bibitem[Xie et~al.(2024)Xie, Lee, Xiao, and Finn]{xie2024decomposing}
Annie Xie, Lisa Lee, Ted Xiao, and Chelsea Finn.
\newblock Decomposing the generalization gap in imitation learning for visual robotic manipulation.
\newblock \emph{International Conference on Robotics and Automation (ICRA)}, 2024.

\bibitem[Yilmaz et~al.(2021)Yilmaz, Hayes, Habib, Burgess, and Barber]{yilmaz2021sample}
Emine Yilmaz, Peter Hayes, Raza Habib, Jordan Burgess, and David Barber.
\newblock Sample efficient model evaluation.
\newblock \emph{arXiv preprint arXiv:2109.12043}, 2021.

\bibitem[Yu et~al.(2020)Yu, Quillen, He, Julian, Hausman, Finn, and Levine]{yu2020meta}
Tianhe Yu, Deirdre Quillen, Zhanpeng He, Ryan Julian, Karol Hausman, Chelsea Finn, and Sergey Levine.
\newblock Meta-world: A benchmark and evaluation for multi-task and meta reinforcement learning.
\newblock \emph{Conference on Robot Learning (CoRL)}, 2020.

\end{thebibliography}

\clearpage
\appendices

\section{Offline Dataset Details}
\label{app:offline_datasets}

\subsection{HAMSTER}
For HAMSTER, we have a cost of 0.5 per execution of an experiment, then an additional switching cost of +1 if a task is of the same task type but requires adding/removing objects.
If a new task type is selected, we then add a cost of +2 for requiring new, often large, objects to be brought into the scene.

\subsection{OpenVLA}
For OpenVLA evaluation, we have a cost of 0.5 per execution of an experiment. 
If a task is changed, such as moving an eggplant to lifting a battery, a cost of 1 is applied.
OpenVLA also has multiple embodiments available, Bridge and the Google Robot.
If there is an embodiment change, we set the changing cost to 3, as this change is relatively large.

\subsection{MetaWorld Policies/Checkpoints}
For MetaWorld evaluation, we have a cost of 0.5 per execution of an experiment.
In MetaWorld tasks, some tasks keep the same objects in the same scene such as opening or closing a window, while others would require new objects like a faucet or a door.
Because these changes are easier to enumerate, we apply only a task switching cost of +1 if the primary object changes, and a switching cost of 0 in the case of the same object being manipulated.

In MetaWorld, we rollout an expert policy for 100 episodes for the 50 tasks to build our training set.
We then train a state-based, language-conditioned behavior cloning policy.
The policy takes in a 768-dimensional language embedding, a 39-dimensional state vector, and outputs a 4-dimensional action. 
For MetaWorld Checkpoints, we train a single MLP-based policy for 100 epochs, recording the policy performance at epoch $1, 10, 20, ..., 100$ for a total of 11 checkpoints. 
For MetaWorld Policies, we instead train 10 policies on random MLP architecture sizes and also apply different amounts of noise to the proprioceptive inputs to the policy to mimic a noisy understanding of state information.
We do this procedure to produce policies that vary more in performance while still having a systematic ``flaw" in understanding the scene, which we hope would be captured in our policy embeddings.
Then, for each policy and environment, we sample 50 evaluations each and store them offline for sampling.

\end{document}